\documentclass[conference]{IEEEtran}
\usepackage{times}

\usepackage[numbers]{natbib}
\usepackage{multicol}
\usepackage[bookmarks=true, unicode, psdextra]{hyperref}

\usepackage{balance}
\usepackage{amsfonts}
\usepackage{amssymb}
\usepackage{lipsum}
\usepackage{siunitx}
\usepackage{booktabs} 
\usepackage{multirow} 
\usepackage{subcaption} 
\usepackage{mathtools}
\usepackage{algpseudocode}
\usepackage{xspace}
\usepackage{algorithm}
\usepackage{rotating}
\usepackage{xcolor}

\newcommand{\arpo}{\gls*{arpo}}
\newcommand{\urlgithub}{\textcolor{magenta}{\url{https://github.com/raphajaner/arpo_racing}}}

\usepackage[nogroupskip,acronyms,nopostdot,style=super,nonumberlist,toc]{glossaries}
\newacronym{arpo}{$\alpha$-RPO}{attenuated residual policy optimization}%
\newacronym{ai}{AI}{artificial intelligence}%
\newacronym{drl}{DRL}{deep reinforcement learning}
\newacronym{dl}{DL}{deep learning}
\newacronym{ml}{ML}{machine learning}
\newacronym{rl}{RL}{reinforcement learning}
\newacronym{ad}{AD}{autonomous driving}
\newacronym{av}{AV}{autonomous vehicle}
\newacronym{nn}{NN}{neural network}
\newacronym{dnn}{DNN}{deep neural network}
\newacronym{ann}{ANN}{artificial neural network}
\newacronym{dqn}{DQN}{deep Q-network}
\newacronym{cnn}{CNN}{convolutional neural network}
\newacronym{rnn}{RNN}{recurrent neural network}
\newacronym{rdqn}{RDQN}{recurrent deep Q-network}
\newacronym{ddqn}{DDQN}{double deep Q-network}
\newacronym{marl}{MARL}{multi-agent reinforcement learning}
\newacronym{dmarl}{DMARL}{deep multi-agent reinforcement learning}
\newacronym{mdp}{MDP}{Markov decision process}
\newacronym{mlp}{MLP}{multilayer perceptron}
\newacronym{mpc}{MPC}{model predictive control}
\newacronym{its}{ITS}{intelligent transportation systems}
\newacronym{ttc}{TTC}{time-to-collision}
\newacronym{vae}{VAE}{variational auto-encoder}
\newacronym{mas}{MAS}{multi-agent system}
\newacronym{mal}{MAL}{multi-agent learning}
\newacronym{per}{PER}{prioritized experience replay}
\newacronym{a2c}{A2C}{advantage actor critic}
\newacronym{sg}{SG}{stochastic game}
\newacronym{mg}{MG}{Markov game}
\newacronym{pomdp}{POMDP}{partially observable Markov decision process}
\newacronym{pomg}{POMG}{partially observable Markov game}
\newacronym{dpomdp}{dec-POMDP}{decentralized partially observable Markov decision process}
\newacronym{nrmse}{NRMSE}{normalized root-mean-square error}
\newacronym{ppo}{PPO}{proximal policy optimization}
\newacronym{gae}{GAE}{generalized advantage estimate}
\newacronym{rpl}{RPL}{residual policy learning}
\newacronym{apf}{APF}{articial potential field}
\newacronym{lstm}{LSTM}{long short-term memory}
\newacronym{ftg}{FTG}{follow-the-gap}
\newacronym{il}{IL}{imitation learning}
\newacronym{rrt}{RRT}{rapidly-exploring random tree}
\newacronym{torcs}{TORCS}{The Open Racing Car Simulator}
\newacronym{cbf}{CBF}{control barrier function}
\newacronym{fov}{FOV}{field-of-view}
\newacronym{de}{DE}{disparity extender}
\newacronym{bc}{BC}{behavior cloning}
\newacronym{ssl}{SSL}{self-supervised learning}
\newacronym{is}{IS}{importance sampling}
\newacronym{poi}{POI}{point-of-interest}

\begin{document}

\title{Efficient Real-World Autonomous Racing via\\Attenuated Residual Policy Optimization}

\author{
\authorblockN{Raphael Trumpp, Denis Hoornaert, Mirco Theile, and Marco Caccamo}
\authorblockA{Technical University of Munich\\
Munich, Germany\\
Email:\{raphael.trumpp, denis.hoornaert, mirco.theile, mcaccamo\}@tum.de}
}


\maketitle

\begin{abstract}
Residual policy learning (RPL), in which a learned policy refines a static base policy using deep reinforcement learning (DRL), has shown strong performance across various robotic applications.
Its effectiveness is particularly evident in autonomous racing, a domain that serves as a challenging benchmark for real-world DRL.
However, deploying RPL-based controllers introduces system complexity and increases inference latency.
We address this by introducing an extension of RPL named attenuated residual policy optimization ($\alpha$-RPO).
Unlike standard RPL, $\alpha$-RPO yields a standalone neural policy by progressively attenuating the base policy, which initially serves to bootstrap learning.
Furthermore, this mechanism enables a form of privileged learning, where the base policy is permitted to use sensor modalities not required for final deployment.
We design $\alpha$-RPO to integrate seamlessly with PPO, ensuring that the attenuated influence of the base controller is dynamically compensated during policy optimization.
We evaluate $\alpha$-RPO by building a framework for 1:10-scaled autonomous racing around it.
In both simulation and zero-shot real-world transfer to Roboracer cars, $\alpha$-RPO not only reduces system complexity but also improves driving performance compared to baselines---demonstrating its practicality for robotic deployment.
Our code is available at: \urlgithub.
\end{abstract}

\IEEEpeerreviewmaketitle

\section{Introduction}\label{sec:introduction}
Once primarily confined to arcade games \cite{mnih2015human}, \gls*{drl} is now being successfully employed across a wide range of real-world domains \cite{tang2025deep}.
However, deployment in real-world autonomous systems introduces new practical challenges, notably in the form of the widely discussed sim-to-real gap and the corresponding complicated task of thorough testing. Prior research has attempted to facilitate learning for such autonomous applications by introducing \gls*{rpl} \cite{silver2018residual, johannink2019residual}.
In this training paradigm, a \gls*{drl} policy is trained to refine the output of a fixed base policy, typically a classical controller, which serves as a strong inductive bias during learning. \gls*{rpl} has been successfully applied to robotic applications such as manipulation \cite{silver2018residual, johannink2019residual, huang2025efficient}, legged locomotion \cite{li2021reinforcement}, drone control \cite{zhang2025proxfly}, and particularly autonomous racing ~\cite{evans2021learning, zhang2022residual, trumpp2023residual, trumpp2024racemop}.

\begin{figure}[t!]
  \centering
  \includegraphics[width=0.475\textwidth]{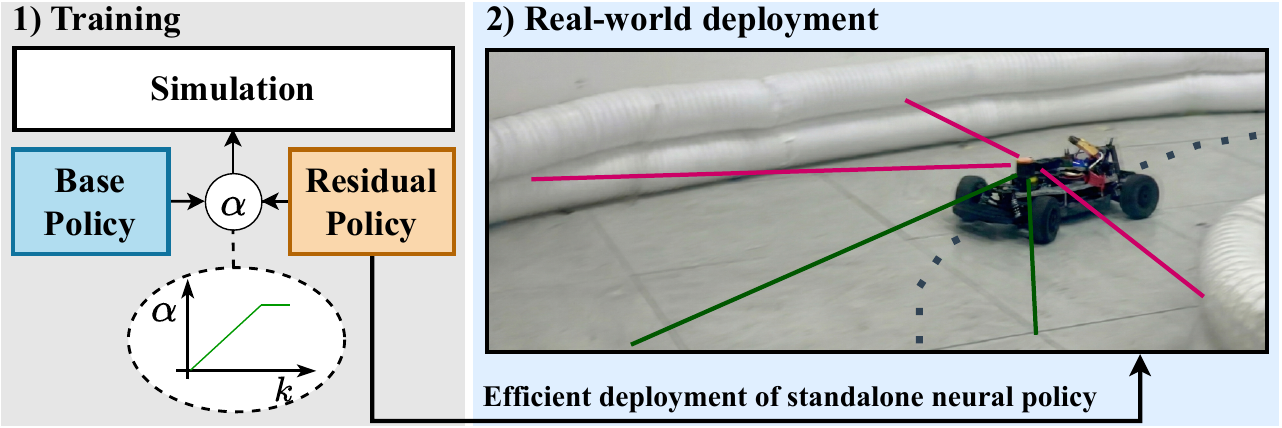} 
  \caption{We test our proposed $\alpha$-RPO method by learning competitive real-world racing behavior with 1:10-scaled autonomous Roboracer cars.
  Compared to classical RPL, $\alpha$-RPO attenuates the contribution of the base policy during training, improving final performance while yielding a standalone neural policy at inference time for efficient deployment.}
  \label{fig:cover}
\end{figure}

As an application, the 1:10-scaled autonomous racing series called Roboracer~\cite{okelly2020f1tenth, evans2024unifying}, formerly known as F1TENTH, has emerged as a prime testbed for learning-based methods as it demands both real-time decision-making and robustness to uncertain conditions while providing well-defined objectives.
Moreover, the ecosystem comes with a broad collection of controllers to compare or use as a base policy for \gls*{rpl}.
As such, a large set of classical racing methods typically relies on offline racing line optimization and subsequent path tracking for deployment~\cite{betz2022autonomous}.
Although successful in Roboracer events~\cite{baumann2025forzaeth}, this approach is not fully autonomous; it depends on pre-computed global trajectories, rendering it unable to adapt to new environments online.
Thus, reactive mapless methods, such as the \gls*{ftg} controller \cite{sezer2012novel}, remain a popular choice for their reliability in real-world Roboracer competitions.

Recent work demonstrated the effectiveness of \gls*{rpl} for autonomous racing in the classical offline case \cite{trumpp2023residual, hildisch2025drive, ghignone2025rlpp} and in the reactive online scenario \cite{zhang2022residual, trumpp2024racemop}; they also include successful real-world deployments \cite{ghignone2025rlpp, hildisch2025drive}.
A remaining primary challenge when using \gls*{rpl} is deciding on the static mixing parameter between the base and residual policies. 
Giving more weight to the residual policy might destabilize early training through poor random decisions.
However, later in training, it may be necessary to allow the agent to override the base policy's decisions.
Popular methods \cite{hildisch2025drive, trumpp2023residual} simply restrict the scale of the residual policy's total contribution to stabilize early training, meaning these works cannot fully explore the agent's potential.
In addition, these \gls*{rpl}-based agents must execute the base policy during deployment, thereby inheriting its specific input modality requirements.
This dependence is particularly prohibitive when the base policy relies on localization, as it increases computation and complicates real-world deployment.

Currently, the prevailing approach in \gls*{rpl} relies on summing the base and residual actions with a static mixing factor.
Only a few studies have explored adaptive aggregation techniques, e.g., \cite{barekatain2021multi, huang2025efficient} present aggregation techniques for \textit{multiple} base policies. 
Similarly, Nakhaei et al.~\cite{nakhaei2024residual} employ an adaptive mixing to bridge the gap between offline and online settings.
However, their method primarily modulates the base policy to mitigate dynamics shifts without removing it.
The full attenuation during online \gls*{rpl} training has not yet been a targeted objective, despite the compelling motivation for it.

Accordingly, we introduce \arpo{}, in which the weight of the base policy is gradually \emph{attenuated to zero} throughout training.
Thus, \arpo{} takes advantage of the early strong inductive bias of the base policy to bootstrap learning while allowing the residual policy to make categorically different decisions at the end of training.
However, an implication of continuously altering the weighing between the base and residual policies is that the environment is non-stationary from the perspective of the residual policy during training.
We mitigate this aspect by integrating \arpo{} directly into \gls*{ppo} \cite{schulman2017proximal}, employing \gls*{ppo}'s importance sampling-based objective to compensate for the non-stationarity induced by the weight attenuation schedule.

Since \arpo{} yields a standalone neural policy at the end of training, real-world deployment is significantly simplified alongside possible fast GPU-based inference.
This inference efficiency mirrors that of lightweight \gls*{bc} architectures like TinyLidarNet~\cite{zarrar2024tinylidarnet}, but \arpo{} optimizes the policy via active interaction to ensure robustness to unseen states and the ability to outperform the base policy.    
Moreover, \arpo{} naturally avoids the common catastrophic forgetting inherent to teacher-student approaches such as \cite{xing2025bootstrapping}, thereby improving stability.
Additionally, \arpo{} stands in contrast to methods such as FastRLAP \cite{stachowicz2023fastrlap}, which are designed specifically for real-world fine-tuning of a base policy; our method instead is designed for zero-shot real-world transfer of a standalone neural policy after training in simulation.

Given the Roboracer platform's suitability as a real-world testbed and the widespread use of \gls*{rpl} in autonomous racing, we benchmark \arpo{} by developing a methodology for learning competitive racing behavior with Roboracer cars.
As shown in Figure~\ref{fig:cover}, we emphasize the autonomy aspect by focusing on the reactive online planning case at deployment and stressing generalization for zero-shot real-world transfer.

Our key contributions are: (1) the introduction of \arpo{}, a novel extension of \gls*{rpl}; (2) which we built upon to propose an autonomous racing framework for Roboracer racing; and (3) our empirical demonstration that \arpo{} provides an effective inductive bias for learning competitive racing behavior with efficient real-world deployment.
We see potential for widespread adoption of \arpo{} in other fields of robotics and facilitate further development by releasing our source code\footnote{Available at: \urlgithub.}.
\section{Background}\label{sec:background}

    \subsection{Deep Reinforcement Learning}\label{subsec:drl}
    Let the stochastic action policy $\pi_{\theta}(a|s)$, parametrized by $\theta$, be used in a \gls*{mdp} for model-free \gls*{drl} with tuple $(\mathcal{S}, \mathcal{A}, \mathrm{P}, \mathrm{R}, \gamma)$.
    The policy is the mapping of observed states $s \in \mathcal{S}$ to a probabilistic space $\mathcal{P}(\mathcal{A})$ over the continuous actions $a \in \mathcal{A}$.
    A scalar reward $r_{t} = \mathrm{R}(s_t, a_t, s_{t+1})$ is obtained when the system transitions to a new state $s_{t+1}$ from $s_t$ with $t$ denoting the timestep.
    The probability of transitioning to $s_{t+1}$ is defined by $\mathrm{P}: \mathcal{S} \times \mathcal{A} \to \mathcal{P}(\mathcal{S})$ and $\gamma\in(0,1]$ is a discount factor. 
    The expected return $\mathrm{V}^{\pi_{\theta}}(s) = \mathbb{E}_{\pi_{\theta}}\left[\sum_{l=0}^{\infty} \gamma^l r_{t+l} \mid s_t=s\right]$ is maximized by the optimal policy $\pi_{\theta}^{*} = \arg \max_{\pi_{\theta}} \mathrm{V}^{\pi_\theta}(s), \,\, \forall s \in \mathcal{S}$.
    
    In \gls*{ppo}~\cite{schulman2017proximal}, the stochastic action policy $\pi_{\theta}(a|s)$ is parametrized by the policy network $f_{\theta}(s)$ with weights $\theta$.
    This \textit{on}-policy \gls*{drl} method updates the weights $\theta_k$ at iteration $k$ with respect to the advantage function $\mathrm{A}^{\pi_{\theta_k}}(s, a) = \mathrm{Q}^{\pi_{\theta_k}}(s,a) - \mathrm{V}^{\pi_{\theta_k}}_{\phi}(s)$.
    The state value $\mathrm{V}^{\pi_{\theta_k}}_{\phi}(s)$ is estimated by the critic network $f_{\phi}(s)$ with weights $\phi$. 
    The (clipped) \gls*{ppo} objective is given by
    \begin{equation}\label{eq:ppo_loss}
        \mathcal L^P(\theta)= \mathbb{E}_{\mathcal{D}_k}\big[ 
        \min \big(  
        \rho_t({\theta}) \mathrm{A}_t,
         \operatorname{clip}\left(\rho_t({\theta}),
         1 - \epsilon, 1 + \epsilon \right) \mathrm{A}_t
        \big) 
        \big],
    \end{equation}   
    where  $A_t\equiv\mathrm{A}^{\tilde\pi_{\theta_k}}(s_t, a_t)$.
    The expectation in ~\eqref{eq:ppo_loss} is taken over the empirical distribution $(s_t, a_t)\sim\mathcal{D}_k$ induced by the behavior policy $\tilde\pi_{\theta_k}$ on the dataset $\mathcal {D}_k=\{\tau_j\}$ of trajectories $\tau_j=\{(s_t, a_t, r_{t},s_{t+1})\}^{T-1}_{t=0}$ with $T$ timesteps.
    Clipping the probability ratio $\rho_t({\theta}) = \frac{\pi_\theta(a_t | s_t)}{\tilde\pi_{\theta_k}(a_t | s_t)}$ with hyperparameter $\epsilon$ increases training stability by limiting the divergence of the updated policy $\pi_{\theta_{k+1}}$ from the behavior policy $\tilde\pi_{\theta_k}$.
    The ratio  $\rho_t$ can be interpreted as an importance sampling term, easing the on-policy requirement to update~\eqref{eq:ppo_loss}.
    
    \subsection{Residual Policy Learning}\label{subsec:rpl}
    \gls*{rpl}~\cite{silver2018residual, johannink2019residual} is a \gls*{drl} paradigm that constructs the action policy $\pi_{\theta}(\cdot\,|\,s)$ by fusing a static base policy $\mu_{\mathrm B}(s)$, typically a classical controller, with a \emph{learned} residual function $f_{\mathrm R,\theta}(s)$ parameterized by a \gls*{dnn}.
    The residual network's weights $\theta$ are optimized using classical \gls*{drl}.
    
    Formally, let the fused action policy $\pi_{\theta}(\cdot\,|\,s)$ be a probability distribution $\mathrm{D}(h_\theta(s); \Lambda)$.
    The distribution parameters $h_\theta(s) \in \mathcal{H}$ are determined by a fusion map $\mathcal{F}$, which combines the base and the residual components according to
    \begin{equation}\label{eq:rpl_fusion}
        \begin{aligned}
            h_\theta(s) \;=\;& \mathcal F \big(\mu_{\mathrm B}(s),\, f_{\mathrm R,\theta}(s);\; \omega\big), \\
            \pi_{\theta}(\cdot\,|\,s) \;:=\;& \mathrm D \big(h_\theta(s);\Lambda\big),
        \end{aligned}
    \end{equation}
    where $\Lambda$ denotes distribution constraints, e.g., action bounds, and $\omega \in [0,1]$ is a \emph{constant} mixing factor.
    
    Several works~\cite{trumpp2023residual, ghignone2025rlpp} instantiate $\mathcal F$ as additive fusion in action space, $a = \mu_{\mathrm B}(s) + \omega \cdot a_{\mathrm R}$, where $a_{\mathrm R}$ is sampled from a residual distribution parameterized by $f_{\mathrm R,\theta}(s)$ or treated deterministically as the direct output.
    Recently, \cite{trumpp2024racemop} showed the advantages of modeling $\mathrm D$ as a Truncated-Gaussian distribution $\mathcal{N}_{[a,b]}(\mu, \sigma^2)$~\cite{burkardt2014truncated}.
    This distribution is a \emph{renormalized} probability density that enforces bounds $[a,b]$ via truncation instead of hard clipping or squashing.
    Unlike non-linear squashing, e.g., Tanh-Gaussian, the Truncated-Gaussian guarantees \emph{local consistency} with the base action: at zero residual mean, the mode of the fused policy $\pi_\theta(\cdot\,|\,s)$ coincides with the base action $\mu_{\mathrm B}(s)$.
    This ensures that the base policies' inductive bias is correctly preserved during early training.
    
    \subsection{Vehicle Model Dynamics with Pacejka Tires}\label{subsec:vehicle_model}
    The state vector of a race car can be modeled as a dynamic single‐track with nonlinear tire forces \cite{ghignone2025rlpp} given by
    \begin{equation}
    \mathbf{x} = \bigl[x \quad y \quad \delta \quad v \quad \psi \quad \dot\psi\quad \beta \bigr]^\mathsf{T},
    \end{equation}
    with the global vehicle position in $x,y$ coordinates and vehicle orientation $\psi$.
    Moreover, the state contains the vehicle's steering angle $\delta$ and driving speed $v$ in addition to the slip angle $\beta$ and yaw rate $\dot\psi$.
    The model input consists of the steering velocity $\dot\delta$ and longitudinal acceleration $a$.
    For the nonlinear tire model, the front and rear tire slip angles are
    \begin{equation}
        \begin{aligned}
        \alpha_f &= \arctan\tfrac{-v_y-\ell_f  \dot\psi}{v_x} + \delta, 
        &
        \alpha_r &= \arctan\tfrac{-v_y+\ell_r  \dot\psi}{v_x},
        \end{aligned}
    \end{equation}
    where $\ell_f,\ell_r$ are the distances from the center of gravity to the front and rear axles.
    Given the normal loads $F_{z,i}$ of the front and rear tires $i \in \{f, r\}$, the corresponding lateral tire forces $F_{y,i}$ are calculated using Pacejka's Magic Formula \cite{pacejka1997magic}
    \begin{align}
        F_{y,i} &= \mu D_i F_{z,i} \sin \Bigl(
          C_i \arctan\!\bigl(
            B_i \alpha_i - E_i [B_i \alpha_i \notag\\
            &\quad - \arctan(B_i \alpha_i)]
          \bigr)
        \Bigr),
    \end{align}
    where $\mu$ is the friction coefficient and $(B_i,C_i,D_i,E_i)$ are tire-specific parameters.
    This model captures realistic driving effects such as understeering and slip in simulation.
    
    \subsection{Autonomous Racing Control}\label{subsec:auto_control}
    Autonomous racing control methods vary in their use of prior information.
    Reactive methods depend solely on onboard sensing, while map-based approaches require an offline optimized racing line and accurate localization at deployment.
    
    The \gls*{ftg} controller~\cite{sezer2012novel} is a reactive LiDAR-based method that identifies the largest collision-free gap in the scan and steers toward its center.
    This logic allows racing without prior maps, but its limited foresight may lead to non-optimal trajectories.  
    The Stanley controller~\cite{thrun2006stanley} is a map-based path-tracking algorithm that computes steering from cross-track and heading errors relative to the reference line.
\section{Attenuated Residual Policy Optimization}\label{sec:arpl}
We propose \arpo{} as a straight extension of \gls*{rpl} that is systematically integrated into \gls*{ppo}.
The core idea of \arpo{} is the fusion mechanism: unlike standard \gls*{rpl}, we progressively \emph{attenuate} the influence of the base policy $\mu_{\mathrm B}(s)$ over the course of training until it is fully removed---only the residual network $f_{\mathrm R, \theta}(s)$ is deployed at inference time as it fully determines the action policy $\pi_\theta(\cdot|\,s)$.
As shown in Algorithm~\ref{alg:arpl}, \arpo{} follows the standard \gls*{ppo} algorithm but leverages the \gls*{ppo} objective to introduce the \emph{synchronization trick} to mitigate consistency issues caused by the attenuation.

In the following, we assume $\mathrm{D}$ belongs to the family of Gaussian distributions, defined by the parameters $(\mu, \sigma) \in \mathcal H$.

    \subsection{Policy Fusion}
    The policy fusion in \arpo{} extends \eqref{eq:rpl_fusion} by introducing the \emph{iteration-dependent} attenuation factor \(\alpha\in[0,1]\)
    \begin{equation}\label{eq:arpl_fusion}
        \begin{aligned}
          h_\theta(s; \alpha) =& \mathcal F\big(\mu_{\mathrm{B}}(s),\, f_{\mathrm R,\theta}(s);\; \alpha\big)\ \\
          \pi_{\theta}(\cdot|s;\alpha) :=& \mathrm D \big(h_\theta(s;\alpha);\Lambda \big),
        \end{aligned}
    \end{equation}
    to control the base policy’s influence on the policy $\pi_{\theta}(\cdot|s;\alpha)$ by defining the distribution's parameters as
    \begin{equation}\label{eq:arpl_params}
        \begin{aligned}
        \mu:=\mu_{\theta}(s) &= (1 - \alpha) \cdot \mu_{\mathrm{B}}(s)  + \max(\alpha, \alpha_{\mathrm{init}}) \cdot f_{\mathrm R,\theta}(s) \\
        \sigma:=\sigma_{\theta}(s) &= \exp(f'_{\mathrm R,\theta}(s)) \quad (f'~\text{uses a separate head).}
        \end{aligned}
    \end{equation}
    The constant coefficient $\alpha_{\mathrm{init}} \in [0,1]$ sets the initial scaling of the residual network.
    We update $\alpha$ using a linear schedule
    \begin{equation}\label{eq:linear_schedule}
        \texttt{update}\_\alpha(k) = \min\left\{1, \tfrac{k}{K_{\text{end}}} \right\},
    \end{equation}
    where $K_{\text{end}}$ denotes the iteration after which the base policy is fully phased out.
    Given $\alpha \approx 0$ at the beginning of training and due to weight initialization, the agent primarily follows the base policy since $\mathbb{E}[f_{\mathrm R, \theta}(s)] \approx 0$.
    As training progresses and $\alpha$ increases towards $1$, the residual network $f_{\text{R},\theta}(s)$ gradually replaces $\mu_{\text{B}}$ until it becomes fully attenuated at $\alpha=1$.
    
    \begin{algorithm}[!t]
        \caption{\arpo{}}
        \label{alg:arpl}
        \begin{algorithmic}[1]
        \For{$k=0,1,\dots$}
          \If{{not} \texttt{sync}} \Comment{{(unsynchronized)}}
            \State Attenuation factor: $\alpha_{k+1} \gets \texttt{update}\_\alpha(k)$
            \State Set behavior policy $\tilde\pi_{\theta_{k}}(\cdot | s) \gets \pi_{\theta_{k}}(\cdot|s; \alpha_{k+1})$
          \Else
            \State Set behavior policy $\tilde\pi_{\theta_{k}}(\cdot | s) \gets \pi_{\theta_{k}}(\cdot|s; \alpha_{k})$
          \EndIf
          \State Collect $\mathcal D_k$ under $\tilde\pi_{\theta_k}$; compute $\hat R_t$ and $A^{\tilde\pi_{\theta_k}}(s_t,a_t)$ 
          \If{\texttt{sync}} \Comment{{(synchronized)}}
            \State Attenuation factor: $\alpha_{k+1} \gets \texttt{update}\_\alpha(k)$
          \EndIf
          \For{epochs/minibatches}
            \State Import. sampl. ratio $\rho_t(\theta;\alpha_{k+1}) \leftarrow \tfrac{\pi_{\theta}(a_t\,|\,s_t;\alpha_{k+1})}{\tilde\pi_{\theta_k}(a_t\,|\,s_t)}$
            \State Update policy by maximizing $\mathcal L^{\mathrm{P}}(\theta;\alpha_{k+1})$
            \State Update critic by minimizing $\mathcal L^{\mathrm{V}}(\phi)$
          \EndFor
          \State Update configuration: $\theta_{k+1} \leftarrow \theta, \phi_{k+1} \leftarrow \phi$
        \EndFor
        \end{algorithmic}
    \end{algorithm}

    \subsection{Synchronization Trick}
    As shown in Algorithm~\ref{alg:arpl}, na\"ively updating $\alpha$ (lines 2--4) alters the behavior policy before data collection.
    Since the policy $\pi_{\theta}(\cdot | s; \alpha)$ is explicitly conditioned on $\alpha$, and the residual network is agnostic to the external modification of the base policy's influence, this creates a \emph{consistency} issue.
    Specifically, if the behavior policy $\tilde\pi_{\theta_k} \gets \pi_{\theta_{k}}(\cdot|s; \alpha_{k+1})$ uses the updated attenuation factor $\alpha_{k+1}$ when interacting with the environment, the sampled actions and probabilities diverge unintentionally from the original policy $\pi_{\theta_k}(\cdot | s; \alpha_{k})$, which resulted from the previous parameter update (line 17).
    
    We mitigate this issue in \arpo{} by proposing the \textit{synchronization trick} (lines 9--11): we use the behavior policy $\tilde\pi_{\theta_k} \gets \pi_{\theta_{k}}(\cdot|s; \alpha_{k})$ to collect rollouts and calculate advantages while keeping $\alpha_k$ constant.
    Only after data collection is $\alpha$ updated to $\alpha_{k+1}$, immediately before the optimization loop.
    This ensures that the shifted influence of the base policy is implicitly absorbed into the importance sampling-based update of $\theta_{k}$ in \gls*{ppo}.
    This relies on the property that, due to the clipped importance sampling ratio $\rho_t$, \gls*{ppo} tolerates mild off-policy updates, i.e., the loss can be computed using rollouts collected from $\tilde\pi_{\theta_k}$ using $\alpha_k$ while optimizing the target policy $\pi_\theta$ using $\alpha_{k+1}$.
    Consequently, this synchronization enables the unbiased use of the current policy for data collection while ensuring correct advantage calculation.

\section{Autonomous Racing Framework}\label{sec:methodology}
In this section, we develop an autonomous racing framework to demonstrate the efficacy of our proposed \arpo{} method.
We build our framework upon the Roboracer platform, a popular real-world autonomous racing series \cite{okelly2020f1tenth, evans2024unifying}. 
Our \arpo{} agent is trained in simulation for the autonomous control of a Roboracer car---with the goal of efficient, zero-shot transfer of the agent to the real-world platform after training. 

    \subsection{Racing Environment}
    
        \subsubsection{Real-World Platform}
        As shown in Figure~\ref{fig:cover}, the Roboracer platform utilizes 1:10-scale remote-control cars, equipped with a 2D LiDAR. 
        This sensor enables the car to perceive its environment through distance measurements; in this case, to the racetrack walls.
        The resulting scan $L_t\in\mathcal{R}^{1081}$ cover a \gls*{fov} of 270$^{\circ}$ and are obtained at 40~Hz.
        The cars are also equipped with an IMU sensor, allowing for direct measurement of the yaw rate $\dot\psi_t$.   
        The longitudinal speed $v^{\text{long}}_{t}$ is estimated based on the motor position.
        The car's control action ${a}=[\delta, v]$ consists of steering angle $\delta$ and target speed $v$ as input to the car's low-level controller.
            
        \subsubsection{Simulation}
        We use a custom version of the Roboracer gym \cite{okelly2020f1tenth} with improved dynamics as simulator, based on the single-track model with the non-linear Pacejka tire model as discussed in Section~\ref{subsec:vehicle_model}.
        The used tire coefficients are realistic estimates, but without real-world identification. 
        We set the maximum vehicle speed to $\SI{8.0}{\meter \per \second}$.
        The observation and control frequencies are fixed to $\SI{40}{\hertz}$, same as the real-world Roboracer's.
        Sensor noise is simulated for the LiDAR scan, speed, and yaw rate measurements as Gaussian noise.

    \subsection{$\alpha$-RPO Agent}
    We adopt~\cite{trumpp2024racemop} and set the policy distribution $\mathrm D$ in \eqref{eq:arpl_fusion} as a Truncated-Gaussian $\mathcal{N}_{[-1,1]}(\mu,\sigma^2)$.
    The base and residual modules of the \arpo{} agent are:
    \begin{itemize}
        \item \textit{Base Module:}
            Our default base module uses the Stanley controller as base policy $\mu_{\text{B}}(s)$ and outputs the base action ${a}_{\mathrm B}=[\delta_{\mathrm B}, v_{\mathrm B}]$.
            When using standard \gls*{rpl}, this controller would require localization during deployment.
            In contrast, \arpo{} leads to a fully \emph{reactive} controller design, as the base policy is fully phased out by the end of training; position information is readily available during training.
            We also ablate using a \gls*{ftg} controller instead. 
            See Section~\ref{subsec:auto_control} for a description of the controllers.
        \item \textit{Residual Module:}
            The residual network $f_{\mathrm R,\theta}(s)$ with weights $\theta$ parametrizes the state-dependent location $\mu_{\theta}(s)$ and scale $\sigma_{\theta}(s)$ according to \eqref{eq:arpl_params}.
    \end{itemize}
    
        \subsubsection{State and Action Space}
        Using only measurements of on-board sensors available on the real-world platform, we define observations $o_t$ at time $t$ as
        \begin{equation}\label{eq:state}
            {o}_{t} = \bigl[L'_t \quad a_{t-1} \quad v^{\text{long}}_{t} \quad \dot\psi_{t} \bigr]^\top,
        \end{equation}
        with the preprocessed LiDAR scan $L'_t$ as the main modality.
        The raw scans $L_t\in\mathcal{R}^{1081}$ are preprocessed by cropping the vector to the central 1,024 readings and normalized to [0, 1].
        The normalized vector is downsampled using average pooling with a kernel size and stride of 2, resulting in the output vector $L'_t \in [0, 1]^{512}$.
        To embed temporal information, we define the current state $s_t \in \mathcal{S}$ as a stack of $n_f=4$ consecutive observations $s_t := (o_{t-n_f+1}, \dots, o_t).$

        After obtaining the fused policy $\pi_{\theta}(\cdot|\,s)$, the target steering and speed are sampled $a_t \sim\pi_{\theta}(\cdot|\,s)$, resulting in bounded values ${a_t}' \in [-1,1]$ due to the used Truncated-Gaussian distribution.
        These values are rescaled to the vehicle's physical action limits to form $a_t \in \mathcal{A}$.

\begin{figure*}[!t]
    \centering
    \begin{subfigure}[b]{0.32\textwidth}
        \centering
        \includegraphics[width=\textwidth]{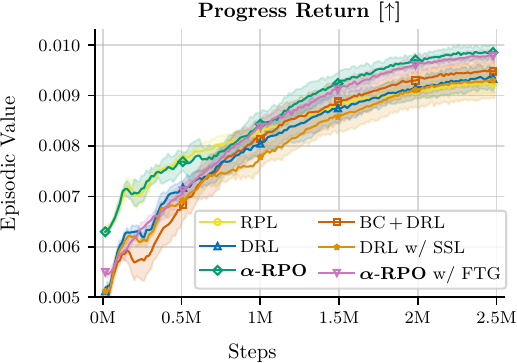}
        \caption{Return corresponding to the reward for lap progress $d^{\text{cl}}_{t+1}$ of \eqref{eq:reward}.}
        \label{fig:train_curve_progress}
    \end{subfigure}
    \hfill
    \begin{subfigure}[b]{0.32\textwidth}
        \centering
        \includegraphics[width=\textwidth]{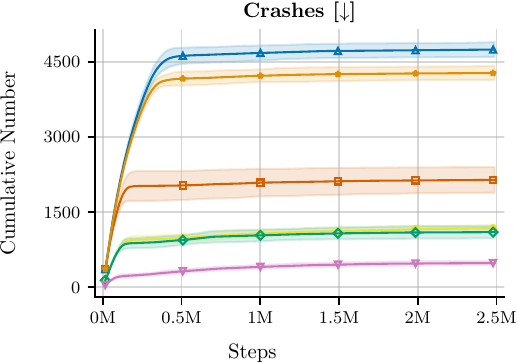}
        \caption{Cumulative number of crashes with the racetrack walls during training.}
        \label{fig:train_curve_collision}
    \end{subfigure}
    \hfill
    \begin{subfigure}[b]{0.32\textwidth}
        \centering
        \includegraphics[width=\textwidth]{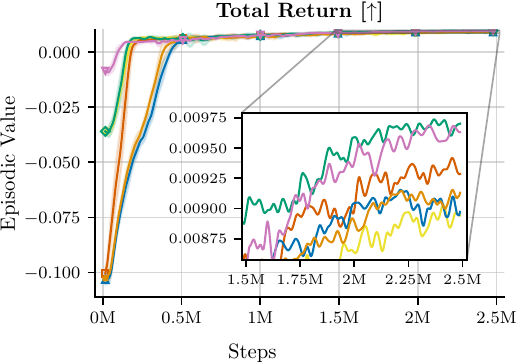}
        \caption{Total return; the zoomed figure shows the last 1M steps in detail.}
        \label{fig:train_curve_totals}
    \end{subfigure}
    \caption{Learning curves with the episodic values of the return during training, showing both the fraction of the return corresponding only to the lap progress (\textbf{left}) and the total sum (\textbf{right}). Additionally, the cumulative number of crashes during training is shown (\textbf{mid}). The agents are trained for 2.5M interaction steps on 15 different maps.}
    \label{fig:training_curves_head_all}
\end{figure*}

        \subsubsection{Reward Formulation}
        We use the progress along the centerline $d^{\text{cl}}_{t}$ as the main optimization target, i.e., the distance between projections of the car's position onto the centerline at subsequent timesteps.
        In addition, our reward function
        \begin{equation}
                 r_{t} =\, 0.1 \cdot d^{\text{cl}}_{t+1} - 0.001 \cdot |\delta_{t+1} - \delta_{t}| - 5 \cdot \mathbf{1}_{c},
             \label{eq:reward}
        \end{equation}
        penalizes large steering action changes between time steps and contains a penalty with $\mathbf{1}_{c}=1$ for collisions with the walls.

        \subsubsection{Attenuation Schedule}
        The attenuation factor $\alpha$ is scheduled using \eqref{eq:linear_schedule} with $K_{\mathrm{end}}$ set to 25\,\% of the total training steps.
        The initial scaling factor is set to $\alpha_{\mathrm{init}}=1$, meaning that the base policy decreases gradually during training while the residual network scaling remains constant.

        \subsubsection{Neural Network Architecture}        
        We employ a \gls*{dnn} architecture comprising a shared encoder with separate heads for the residual network $f_{\mathrm{R}, \theta}$ and critic network $f_{\phi}$, totaling 344,357 trainable parameters.
        The LiDAR encoder processes $L'_t$ by a stack of five 1D convolutional layers.
        To extract a compact feature representation from the convolutional feature maps, we apply a SpatialSoftmax layer \cite{finn2016deep} followed by Linear projection, yielding the final LiDAR embedding ${z}_t \in \mathbb{R}^{128}$.
        The embedding ${z}_t$ is concatenated with the encoded vehicle states $h_t$ and normalized via LayerNorm in the Fusion Block.
        This fused feature vector serves as the input to the independent fully connected layers of the actor and critic heads.
        Appendix~\ref{app:nn_design} contains a visualization of this \gls*{dnn}.
\section{Experiments}\label{sec:experiments}

We use the presented autonomous racing framework to evaluate the performance of our proposed \arpo{} method.
We first discuss simulation results, as this allows a detailed comparison to other methods.
Eventually, we zero-shot transfer the agents trained with \arpo{} to the real world and conduct a study on a real-world racetrack using a Roboracer car.
    
        \subsubsection{Training}
        We run training for 2.5\,M timesteps, using 15 parallel environments to collect trajectories of 1024 timesteps.   
        Due to the high efficiency of the methodology and parallelized training, full training takes \emph{less than 15\,min} on a workstation with an AMD Ryzen 9 9950X CPU and an NVIDIA GeForce RTX 5090 GPU.
        We ensure reproducibility by listing simulation parameters and hyperparameters of \gls*{ppo} in Appendix~\ref{app:reproducability}.
              
        \subsubsection{Racetracks}
        Our pipeline uses 15 synthetically generated racetracks during training.
        Additionally, another 6 racetracks are used for testing generalization: 5 are also synthetically generated, while the \emph{Munich} racetrack is the SLAM-based map of our own \emph{real-world} racetrack.
        Renderings of all racetracks are presented in Appendix \ref{app:racetracks}.
        
        \subsubsection{Evaluation}
        We evaluate on against-the-clock races that finish after completion of \emph{5 laps}.
        Like in the Roboracer competitions, if a crash happens during evaluation, (1) the race is stopped and the car is relocated to the nearest point on the centerline, (2) a 5-second penalty is added to the lap time, and (3) the race resumes.
        Per-racetrack results are averaged over 10 different starting positions on the track.
        Unless otherwise stated, our presented results are mean values from 5 independent training runs with different seeds, and shaded areas show the corresponding standard deviation.
        
        \subsubsection{Baselines}
        To provide comprehensive results, we include results from multiple related learning-based baseline methods: (1) standard \gls*{rpl}, (2) standard \gls*{drl}, (3) \gls*{drl} with \gls*{bc}-based pretraining of the actor using a recorded dataset of Stanley-based trajectories of 0.5\,M steps, and (4) \gls*{drl} with an additional \gls*{ssl} for distilling the base action during training.
        All these methods use \gls*{ppo} with the same configuration as \arpo{}.
        Additionally, the two classical controllers, (5) \gls*{ftg} and (6) Stanley, are evaluated to provide a better perspective on the learning-based methods.
        Both controllers use a single set of parameters optimized for robustness to prevent collisions across all training racetracks, which may result in increased lap times on individual tracks.
    
    \subsection{Benchmark}
            
    \begin{table*}[!t]%
        \centering%
        \caption{Simulated racing results, showing a subset of 7 training and 6 testing racetracks. The \emph{Munich} racetrack is an SLAM-based map of our own real-world racetrack. A race finishes when \textit{5 laps} are completed. The acronym BCD is short for the BC+DRL method. Full results including all maps and DRL+SSL are shown in Appendix~\ref{app:extended_experiments}.}
        \label{tab:lap_results}
        \setlength{\tabcolsep}{1mm}%
        \begin{tabular}{p{0.2cm} p{1.7cm} | cccccc | cccccc | cccccc}%
        \toprule%
        \multicolumn{2}{c |}{\multirow{2}{*}{\textbf{Racetrack}}}&\multicolumn{6}{c |}{\textbf{Total race time} $[\downarrow]$}&\multicolumn{6}{c |}{\textbf{Collisions per lap} $[\downarrow]$}&\multicolumn{6}{c}{\textbf{Maximum speed} $[\uparrow]$}\\%
        &&FTG&Stanley&DRL&BCD&RPL&\underline{$\alpha$-RPO}&FTG&Stanley&DRL&BCD&RPL&\underline{$\alpha$-RPO}&FTG&Stanley&DRL&BCD&RPL&\underline{$\alpha$-RPO}\\%
        \midrule%
        \multirow{6}{*}{\rotatebox[origin=c]{90}{\textbf{Train}}}
        &Abu Dhabi&67.54&50.41&48.50&46.29&48.16&44.02&0.00&0.00&0.03&0.00&0.09&0.00&4.32&5.06&4.81&4.72&5.15&5.15\\%
        &London&71.04&55.72&55.73&50.05&52.57&47.58&0.00&0.00&0.20&0.00&0.04&0.00&5.03&4.81&5.00&4.84&5.10&5.29\\%
        &Mexico City&55.78&41.15&37.51&37.54&37.78&35.53&0.04&0.00&0.00&0.00&0.00&0.00&4.78&5.21&5.10&4.92&5.37&5.50\\%
        &New York&71.88&63.08&54.49&54.00&56.17&51.95&0.00&0.00&0.00&0.00&0.00&0.00&4.58&4.45&5.08&4.95&4.89&5.34\\%
        &Paris&51.40&39.74&41.08&35.88&35.78&34.13&0.00&0.00&0.19&0.00&0.00&0.00&4.48&5.32&5.25&5.00&5.41&5.50\\%
        &Sydney&55.95&44.52&40.18&40.16&40.43&38.37&0.00&0.00&0.00&0.00&0.00&0.00&4.94&4.98&5.23&4.91&5.38&5.44\\%
        &$\dots$\\
        \midrule%
        &Average Train&67.74&53.85&49.92&48.31&49.84&\textbf{46.24}&0.01&\textbf{0.00}&0.04&{0.00}&0.01&\textbf{0.00}&4.70&5.00&5.10&4.93&5.17&\textbf{5.41}\\%
        \midrule%
        \multirow{7}{*}{\rotatebox[origin=c]{90}{\textbf{Test}}}
        &Buenos Aires&67.29&52.33&49.79&49.09&50.30&47.30&0.00&0.00&0.00&0.00&0.00&0.00&4.40&4.85&4.90&4.86&4.88&5.24\\%
        &Istanbul&69.62&52.27&48.15&53.01&86.42&45.86&0.00&0.00&0.00&0.19&1.24&0.00&4.48&5.32&4.95&4.82&5.30&5.29\\%
        &Madrid&66.91&53.51&49.66&49.09&50.87&47.28&0.00&0.00&0.00&0.00&0.00&0.00&4.97&4.46&5.03&4.88&4.79&5.32\\%
        &Seoul&77.78&65.95&61.57&59.96&65.19&57.76&0.00&0.00&0.00&0.00&0.08&0.00&4.94&5.03&5.21&5.00&5.10&5.43\\%
        &Toronto&76.31&65.20&59.63&57.92&62.67&55.58&0.00&0.00&0.00&0.00&0.00&0.00&5.18&4.60&5.16&4.95&4.95&5.35\\%
        &\emph{Munich}&38.10&32.77&41.87&33.60&32.71&28.91&0.00&0.00&0.41&0.14&0.01&0.00&3.41&4.29&4.69&4.70&4.46&4.92\\%
        \midrule%
        &Average Test&66.00&53.67&51.78&50.44&58.03&\textbf{47.11}&\textbf{0.00}&\textbf{0.00}&0.07&0.06&0.22&\textbf{0.00}&4.57&4.76&4.99&4.87&4.92&\textbf{5.26}\\%
        \bottomrule%
        \end{tabular}%
    \end{table*}

    \begin{figure*}[!t]
        \centering
        \begin{subfigure}[b]{0.32\textwidth}
            \centering
            \includegraphics[width=\textwidth]{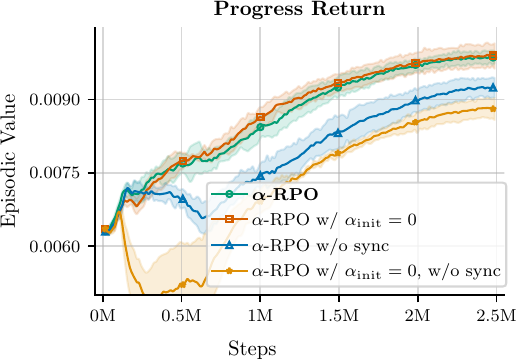}
            \caption{Ablation on the \textbf{synchronization trick}.}
            \label{fig:ablation_sync_trick}
        \end{subfigure}
        \hfill
        \begin{subfigure}[b]{0.32\textwidth}
            \centering
            \includegraphics[width=\textwidth]{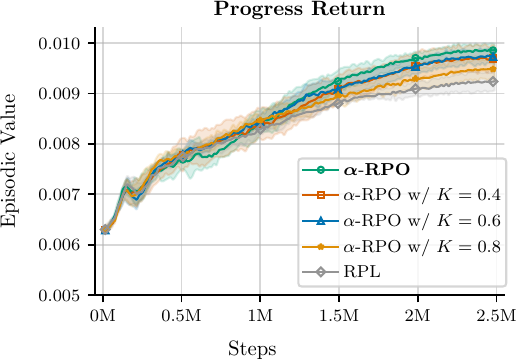}
            \caption{Ablation on the  $\boldsymbol\alpha$ \textbf{schedule}.}
            \label{fig:ablation_schedule}
        \end{subfigure}
        \hfill
        \begin{subfigure}[b]{0.32\textwidth}
            \centering
            \includegraphics[width=\textwidth]{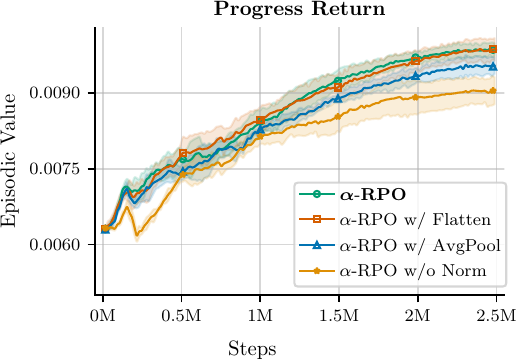}
            \caption{Ablation on the \textbf{DNN design}.}
            \label{fig:ablation_dnn_design}
        \end{subfigure}
        \caption{Learning curves showing the progress-return for our ablation studies. We ablate the synchronization trick (\textbf{left}), longer $\alpha$-schedules (\textbf{mid}), and design aspects of the DNN (\textbf{right}).}
        \label{fig:training_curves_ablations}
    \end{figure*}
    
        \subsubsection{Training Curves}
        We evaluate the training curves of \arpo{} in comparison to the baselines, visualizing first the reward corresponding to the \emph{progress} $d^{\text{cl}}_{t+1}$ of \eqref{eq:reward} as this is the best indicator of how fast the agents have learned to race.
        As expected, Figure~\ref{fig:train_curve_progress} shows that using a base policy is a strong inductive bias that bootstraps early learning in \arpo{} and \gls*{rpl} as the vehicle \textit{can drive} directly when learning starts.
        We see that \arpo{} and \gls*{rpl} initially perform very similarly for 0.6\,M steps.
        After a short phase of instability, \arpo{} gains an edge in performance after approx. 0.8\,M steps, eventually achieving the highest average progress of 0.99.
        This result supports the hypothesis that the base policy is only a meaningful inductive bias in early training but hinders optimal final performance, as \gls*{rpl} agents must continuously override the base policy's behavior.
        As a result, after full training, \gls*{drl} and \gls*{rpl} perform almost identically\footnote{It is noteworthy that \gls*{drl} itself also starts with non-zero progress as an inductive bias arises from the action space definition itself, leading to an expected speed target of $\SI{4}{\meter\per\second}$ at initialization.}.

        Additionally, our results show that \gls*{drl}-based methods can outperform \gls*{rpl} when using \gls*{bc}-based pretraining for the agent or adding a \gls*{ssl} loss during training.
        This reiterates our aforementioned hypothesis that using the base policy in \gls*{rpl} is primarily useful during early training but may lead to lower final performance.
        The additional result of \arpo{} w/ \gls*{ftg} demonstrates that, while \arpo{} benefits from the faster Stanley base policy, the learning paradigm provides a general improvement over all baselines.
        The results in Figure~\ref{fig:train_curve_collision}, which show the cumulative number of crashes during training, confirm this aspect since the \gls*{drl}-based methods exhibit a substantially higher collision count during training.
        In contrast, both \gls*{rpl} and \arpo{} sustain less than 1,500 total collisions during training; \gls*{bc}-pretraining is also effective at reducing the large amount of collisions during the early learning phase.
        Notably, the \arpo{} agent using the \gls*{ftg} base policy causes fewer than 500 total crashes, likely since the Stanley already operates at a higher racing speed during exploration.
        
        In general, it can be seen that \arpo{} leverages the advantages of \gls*{rpl} during early training but is capable of higher final performance after attenuation of the base policy. 
        As such, we understand \arpo{} as a form of guided exploration which helps to bootstrap \gls*{drl} learning.

        \subsubsection{Lap Results}           
        Table~\ref{tab:lap_results} summarizes the performance of \arpo{} against the baselines.
        \arpo{} consistently outperforms all other methods, achieving the lowest average total race time of $\SI{46.24}{\second}$ on training and $\SI{47.11}{\second}$ on unseen test tracks.
        Notably, \arpo{} demonstrates superior generalization capabilities compared to standard \gls*{rpl}.
        While standard \gls*{rpl} degrades significantly on unseen tracks to an average of $\SI{58.03}{\second}$ with $\SI{0.22}{}$ collisions per lap, \arpo{} maintains robust performance with zero collisions.
        Furthermore, the maximum velocity indicates that \arpo{} races closer to the physical limit, reaching the highest average max speed of $\SI{5.41}{\meter\per\second}$.
        While the BC+DRL method has learned to prevent all collisions during training, the agent shows lower generalization capabilities than \arpo{}.
        On our own real-world-based racetrack, \emph{Munich}, \arpo{} improves lap times by over $\SI{12}{\percent}$ compared to Stanley, indicating strong potential for effective sim-to-real transfer. 
        
        \begin{figure*}[!t]
            \centering
            \begin{subfigure}[b]{0.325\textwidth}
                \centering
                \includegraphics[width=1.0\linewidth]{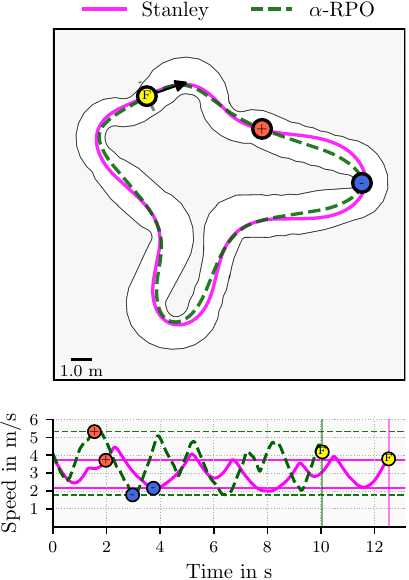}
                \caption{Stanley (magenta) against \arpo{} (green) on the New York racetrack (training).}
                \label{fig:trajectory_1}
            \end{subfigure}
            \hfill
            \begin{subfigure}[b]{0.325\textwidth}
                \centering
                \includegraphics[width=\linewidth]{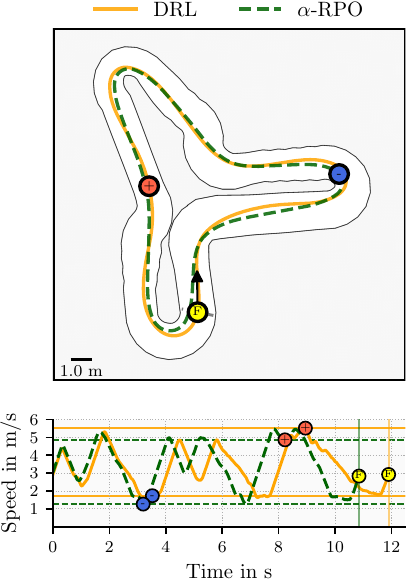}
                \caption{\gls*{drl} (orange) against \arpo{} (green) on the Toronto racetrack (testing).}
                \label{fig:trajectory_2}
            \end{subfigure}
            \hfill
            \begin{subfigure}[b]{0.325\textwidth}
                \centering
                \includegraphics[width=0.9\linewidth]{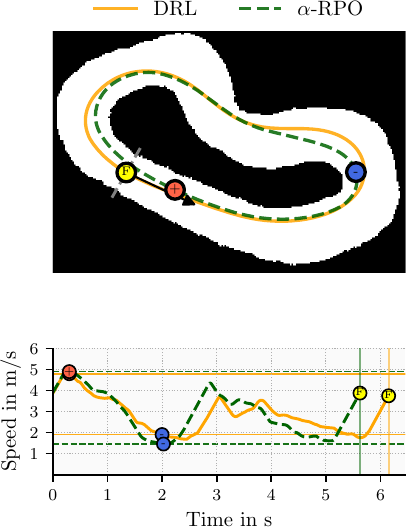}
                \caption{\gls*{drl} (orange) against \arpo{} (green) on the Munich racetrack (testing).}
                \label{fig:trajectory_3}
            \end{subfigure}
            \caption{Qualitative comparison of trajectories (\textbf{top}) and speed profiles (\textbf{bottom}) on three racetracks, showing a single flying lap. Markers indicate a specific \gls*{poi} on the racetracks: $+$ marks a high-speed section (red), $-$ indicates the section with minimal speed (blue), while $\mathrm F$ is the finish line (yellow).} 
            \label{fig:trajectory_analysis}
        \end{figure*}

    \subsection{Ablation Studies}\label{subsubsec:ablations}
    We validate our design choices through detailed ablations, focusing on the progress return as performance measure.
    \begin{itemize}
        \item Figure~\ref{fig:ablation_sync_trick} confirms the necessity of our \emph{synchronization trick}; it stabilizes training and is a key contribution of \arpo{}.
                The effectiveness of leveraging the \gls*{is} term in \eqref{eq:ppo_loss} to reduce the bias when changing the attenuation factor is even more apparent when setting $\alpha_{\mathrm{init}}=0$. Here, the residual network's scale also changes continuously during the attenuation phase, leading to substantial instability without synchronization. 
        \item Figure~\ref{fig:ablation_schedule} demonstrates that \emph{shorter} attenuation schedules are advantageous: the longer the attenuation, the closer the agent's performance to standard \gls*{rpl}.
        \item The architectural analysis in Figure~\ref{fig:ablation_dnn_design} identifies the fusion block's LayerNorm for feature normalization as a vital component. Using a SpatialSoftmax layer \cite{finn2016deep} also has advantages: it reduces the feature space and total number of parameters compared to Flatten, but performs better than AveragePooling \cite{trumpp2025impoola} for feature reduction, presumably due to the centered nature of the observation.
    \end{itemize}

    \subsection{Qualitative Analysis of Trajectories}
    To better understand the performance gains, we analyze the resulting trajectories and speed profiles on three representative maps in the following.
    It is apparent from Figure~\ref{fig:trajectory_1} that the racing line used for the Stanley controller is based on minimum-curvature optimization.
    The learned trajectory of \arpo{} follows a similar shape; however, the path length is substantially shortened in several sections by following the inner wall more closely for a straighter path.
    As seen at $+$ marked section, this behavior allows the agent to achieve a higher maximum speed exceeding $\SI{5}{\meter\per\second}$.
    
    Figure~\ref{fig:trajectory_2} and \ref{fig:trajectory_3} compare \arpo{} against the \gls*{drl} baseline on unseen test tracks.
    On the \emph{Toronto} racetrack, \gls*{drl} cannot find the line that cuts corners as aggressively as \arpo{} does.
    The speed profile shows that \arpo{} can accelerate to higher speeds while braking harder, resulting in the significant lap-time reduction.
    Similar behavior is evident on the \emph{Munich} track, where \arpo{} brakes harder at the curve with minimum speed, which allows the agent to turn in harder, thus having a straighter line to accelerate out of the corner.

    \begin{table}[!t]
        \centering
        \caption{Results for fine-tuned agents of the single best training seed, denoted by~\textsuperscript{*}. Fine-tuning continues training only on the specific racetrack for a further 0.5\,M interaction steps.}
        \label{tab:lap_time_finetuned}
        \begin{tabular}{l | r r r r}
        \toprule
            \multirow{2}{*}{\textbf{Racetrack}} & \multicolumn{4}{c}{\textbf{Total race time in s} [$\downarrow$]} \\
            & {DRL} & DRL\textsuperscript{*} & \underline\arpo{}& \underline\arpo{}\textsuperscript{*} \\
            \midrule
            Toronto & 59.95 & 59.37 & 54.90 & \textbf{52.49}\\%
            Madrid & 50.30 & 47.40 & 46.59 & \textbf{43.41} \\%
            Munich & 31.23 & 29.43 & 28.52 & \textbf{26.35} \\%
        \bottomrule
        \end{tabular}
    \end{table}
       
    \subsection{Fine-Tuning on Single Racetracks}
    We explore the capabilities of \arpo{} agents to fine-tune on a single new racetrack after initial training focused on generalization across the 15 training maps.
    Such fine-tuning is highly relevant to real-world racing in Roboracer competitions, as it allows for \emph{pre-training} an agent that is then fine-tuned once the competition's racetrack is accessible. 
    We conduct the fine-tuning by continuing training for 0.5M interaction steps on a single racetrack\footnote{We achieved the best results when reducing the initial learning rate to $5e^{-5}$and collecting data from only 4 parallel environments.}.
    The results in Table~\ref{tab:lap_time_finetuned} substantiate that single-track fine-tuning on new maps leads to a meaningful performance increase; however, \gls*{drl} still lags behind \arpo{}.

    \begin{figure*}[!t]
        \centering
        \begin{subfigure}[t]{0.49\textwidth}
          \centering
          \includegraphics[width=1.0\textwidth]{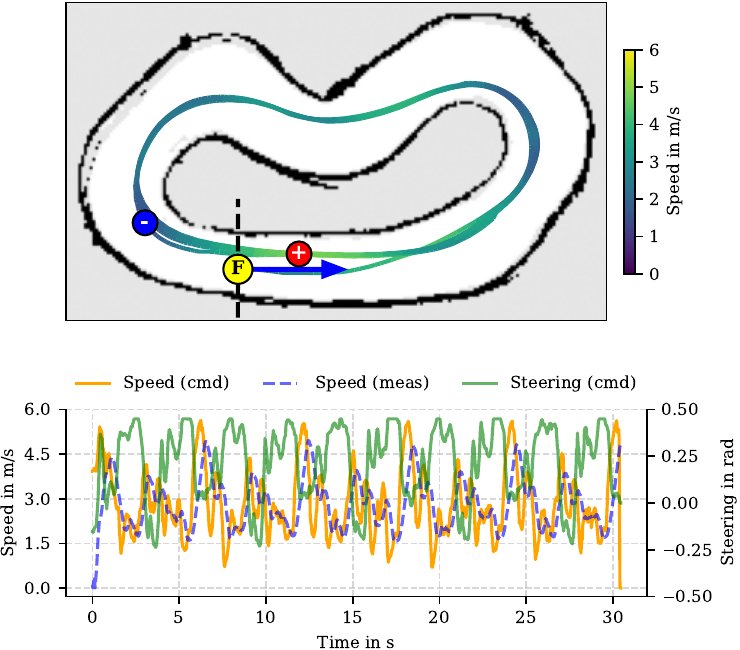}
          \caption{Counterclockwise direction, without obstacle.}
          \label{fig:trajectory_arpl}
        \end{subfigure}
        \hfill
        \begin{subfigure}[t]{0.49\textwidth}
          \centering
          \includegraphics[width=1.0\textwidth]{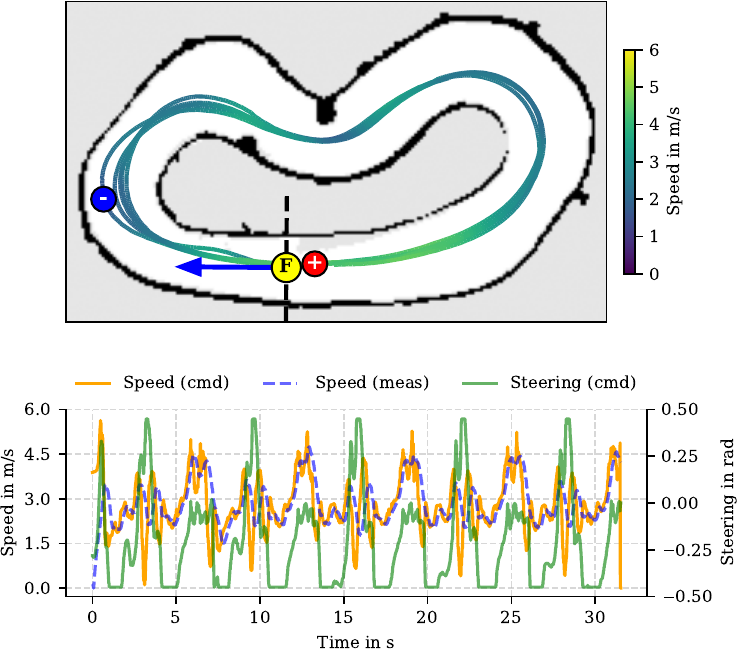}
          \caption{\emph{Clockwise} direction with two \emph{obstacles} placed in the top section.}
          \label{fig:trajectory_arpl_obstacle}
        \end{subfigure}
        \caption{Real-world trajectories of \arpo{} on the \emph{Munich} racetrack for 5 laps (\textbf{top}), including a varaition with two obstacles placed; current speed shown as color bar. Positions are estimated offline from \texttt{rosbag} data using SLAM. The agent's actions command (cmd) and measured speed (meas.) are also shown (\textbf{bottom}). Markers indicate a specific \gls*{poi} on the racetracks: $+$ marks a high-speed section (red), $-$ indicates the section with minimal speed (blue), while $\mathrm F$ is the finish line (yellow).}
    \end{figure*}

    \subsection{Real World Experiments}
        \subsubsection{Deployment}
        The design of \arpo{} offers significant advantages for real-world deployment, as only a standalone \gls*{dnn} is required.
        This case allows for very low inference times~\cite{ghignone2024tcdriver, zarrar2024tinylidarnet} and \emph{removes} the need for complex system integration with the base policy.
        Our Roboracer's onboard embedded board is an NVIDIA Jetson Orin Nano Super, which enables efficient \gls*{dnn} acceleration via GPU cores.
        During deployment of our \arpo{} agent on the Roboracer car, we measure average inference latencies of 3.5~\unit{ms}, including the \emph{full} LiDAR preprocessing and ROS2-based topic publishing stack.
        In contrast, a well-optimized localization-based stack is reported by Baumann et al.~\cite{baumann2025forzaeth} with an average of 7.5~\unit{ms}, even if running on an Intel i5-10210U.
        Moreover, \arpo{} enables a form of \emph{privileged} learning since we do not need to deploy the Stanley controller that was used as base policy; \gls*{rpl} and hybrid approaches~\cite{ghignone2024tcdriver} would still include heavy computational overhead due to the required localization.

        \subsubsection{Zero-Shot Real-World Transfer}
        We visualize recorded real-world trajectories in Figure~\ref{fig:trajectory_arpl}, where we tested an agent on our own \emph{Munich} racetrack.
        Although \emph{Munich} was not used during training, our \arpo{} successfully bridges the sim-to-real gap and generalizes to this new racetrack.
        It can be seen that the agent consistently chooses the same trajectory, following the inside wall closely.
        The measured \emph{real-world} racing results are shown in Table~\ref{tab:real_world_lap_times}.
        Compared to simulated results, \arpo{} achieves a similarly high maximum speed of approx. $\SI{5.0}{\meter\per\second}$ during real-world deployment, but the average single lap duration increases by approx. $\SI{0.4}{\second}$.
        
        We stress the robustness of \arpo{} in Figure~\ref{fig:trajectory_arpl_obstacle}, placing two static \emph{obstacles} on the racetrack.
        The agent adjusts its trajectory smoothly to avoid a collision with the obstacle.
        While these results demonstrate robustness of the \arpo{} agent, they indicate a certain degree of simulation mismatch; for example, there is motor stuttering during the start-up in the first $\SI{1.5}{\second}$, requiring the agent to brake hard in this corner.
        In general, the figures highlight that driving behavior and collision avoidance are consistent and repeatable.
        Please find the video material of these experiments in the Supplementary.

        \begin{table}[!t]
            \centering
            \caption{Real-world lap times on the Munich track, using the model corresponding to the best seed of each method. Finish times cover races with 5 full laps. Reported values are estimated offline from \texttt{rosbag} data.}
            \label{tab:real_world_lap_times}
            \begin{tabular}{l | r r r | r}
              \toprule
              \multirow{2}{*}{\textbf{Metric}} & \multicolumn{4}{c}{\textbf{Method}} \\
               & FTG & DRL & \underline\arpo{} & \underline\arpo{}\textsuperscript{*}\\
              \midrule
              Finishing time in \unit{\second} [$\downarrow$]& 40.2 & 34.4 & \textbf{30.5} & 28.1\\  
              Difference to simulation in \unit{\second} [$\downarrow$]& +2.1 & +3.2 & \textbf{+2.0} & +1.7 \\
              Maximum speed in \unit{\meter\per\second}[$\uparrow$]& 3.1 & 4.8 & \textbf{5.0} & 5.4 \\
            
              \bottomrule
            \end{tabular}
        \end{table}
\section{Limitations and Future Work}
Unlike \gls*{rpl} approaches that use a base policy to provide safety guarantees \cite{cao2024physicsregulated}, \arpo{} yields a non-verifiable neural policy.
While our results support this standalone deployment, the agent forfeits the explicit, verifiable constraints inherent in classical control, hindering formal verification.
Additionally, our simulation parameters were empirically calibrated rather than validated via rigorous system identification.
Although zero-shot transfer was successful, unmodeled dynamic effects seemingly still hinder performance at the limits of handling.
Since our fine-tuning experiment assessed strong effectiveness, we envision an online, real-world fine-tuning procedure for future work.
Furthermore, \arpo{} may remain sensitive to the quality of the base policy and the linear attenuation schedule of $\alpha$. 
An ablation should aim to better understand this relation by testing various base policies and schedules.
Eventually, future work should extend the usage of \arpo{} to other robotic control domains to assess its broader applicability.

\section{Conclusion}\label{sec:conclusion}
This work introduces \arpo{}, an extension of the popular \gls*{rpl} training paradigm.
In contrast to classical \gls*{rpl}, our proposed method progressively attenuates the base policy during training, thereby improving performance by removing the ambiguity between the base and residual policies.
We build an autonomous racing methodology upon \arpo{} for 1:10-scaled autonomous racing to evaluate \arpo{} under challenging conditions.
In both simulation and zero-shot real-world transfer to Roboracer cars, \arpo{} achieves robust, competitive racing behavior.
We confirm that our proposed synchronization trick substantially stabilizes training.
Moreover, our results demonstrate \arpo{}'s practicality for real-world robotic deployment as \arpo{} yields a standalone neural policy after training, which enables simplified, efficient real-world deployment with low inference time.

\section*{Acknowledgments}
Marco Caccamo was supported by an Alexander von Humboldt Professorship endowed by the German Federal Ministry of Education and Research.

\balance
\bibliographystyle{plainnat}
\bibliography{references}

\clearpage
\onecolumn

\appendices

\section{Reproducability}\label{app:reproducability}

    \subsection{Training Hyperparameter}\label{app:ppo_params}
    
    \begin{table}[!h]
        \centering
                \caption{Hyperparameters for PPO backend as used in AR-PO, RPL, and DRL.}
        \label{tab:hyperparameters}
        \setlength{\tabcolsep}{1mm}
        \begin{tabular}{l r | l r }
            \toprule
            \multicolumn{4}{c}{\textbf{Training Hyperparameter}}\\
            \midrule
            Total steps & 2.5e\textsuperscript{6} & Discount $\gamma$ & 0.99 \\ 
            Learning rate (LR) &  1.5e\textsuperscript{-4} & GAE $\lambda$ & 0.95 \\
            LR schedule & $\cos$ &  Value coefficient & 0.5 \\
            Trajectory length & 1024 &  Max. gradient norm & 0.5 \\
            Clip $\epsilon$ & 0.2 & Update epochs & 5 \\
            Batch size & 512 & Initial $\log \sigma_{R,\theta}$ & -1.0 \\
            Initial entropy & 0.01 & Final entropy & 0.0 \\
            Entropy decay & linear  \\
            \bottomrule
        \end{tabular}
    \end{table}

    \subsection{Simulation Parameters}\label{app:sim_params}

        \subsubsection{Vehicle Model}
        Parameters for the vehicle model in Table~\ref{tab:vehicle_params} are estimated from real-world Roboracer cars.
        \begin{table}[!h]
            \centering
            \caption{Simulation parameters for the single-track vehicle model.}
            \label{tab:vehicle_params}
            \setlength{\tabcolsep}{1mm}
            \begin{tabular}{l r | l r }
                \toprule
                \multicolumn{4}{c}{\textbf{Vehicle Parameter}}\\
                \midrule
                Vehicle mass in \unit{\kilogram} & 3.3 & Inertia in \unit{\kilogram\square\meter} & 0.0627 \\
                Vehicle width in \unit{\meter} & 0.27 & Vehicle length in \unit{\meter}& 0.51 \\
                Distance front axle in \unit{\meter} & 0.1625 & Distance rear axle in \unit{\meter} & 0.1625 \\
                CG height in \unit{\meter} & 0.02 & Steering angle in \unit{\radian} & $\pm 0.45$ \\
                Max. velocity in \unit{\meter\per\second} & 8.0 & Min. velocity in \unit{\meter\per\second} & 0.0 \\
                Max. acceleration in \unit{\meter\per\square\second} & $4.9$ & Max. braking in \unit{\meter\per\square\second} & -$3.7$ \\
                Steering velocity in \unit{\radian\per\second} & $\pm 3.2$ & Lidar range in \unit{\meter} & 10.0\\  
                Lidar points & 1081 & Lidar FOV in \unit{\degree}& 270 \\
                LiDAR frequency in \unit{\hertz} & 40 & Control frequency in \unit{\hertz} & 40 \\
                \bottomrule
            \end{tabular}
        \end{table}

        \subsubsection{Tire Model}
        \citet{pacejka1997magic}'s Magic Tire formula uses four coefficients for capturing nonlinear tire behavior.
        The parameter $B$ controls the stiffness, $C$ shapes the curve, $D$ scales the peak force (normalized), and $E$ governs the transition from linear to saturation behavior.
        The friction coefficient $\mu$ scales the forces accordingly.
        Given the vehicle parameters and corresponding normal forces, a rendering of the lateral tire forces for the coefficients in Table~\ref{tab:pacejka_params} is shown in Figure~\ref{fig:tire_forces}.

        \begin{table}[h]
            \centering
            \begin{minipage}{0.48\textwidth}
                \centering
                \caption{Pacejka tire model parameters.}
                \label{tab:pacejka_params}
                \setlength{\tabcolsep}{1mm}
                \begin{tabular}{l r | l r }
                    \toprule
                    \multicolumn{4}{c}{\textbf{Parameter Values}}\\
                    \midrule
                    $\mu$ & 0.5 & & \\
                    $B_f$ & 5.5 & $B_r$ & 6.5 \\
                    $C_f$ & 1.5 & $C_r$ & 1.5 \\
                    $D_f$ & 1.0 & $D_r$ & 1.0 \\
                    $E_f$ & 0.1 & $E_r$ & 0.1 \\
                    \bottomrule
                \end{tabular}
            \end{minipage}%
                \hfill
            \begin{minipage}{0.48\textwidth}
                \centering
                \includegraphics[width=\textwidth]{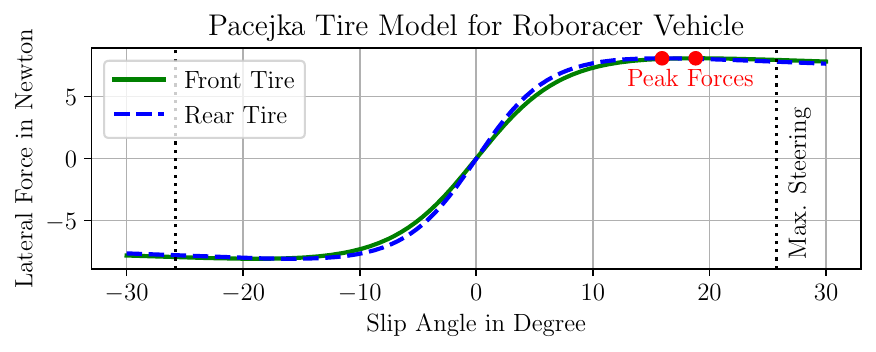}
                \captionof{figure}{Tire lateral forces over slip angle using Pacejka model.}
                \label{fig:tire_forces}
            \end{minipage}
        \end{table}

    \subsection{Software}
    We list critical software with versions and used open-source repositories here:
    \begin{itemize}
        \item Python packages
        \begin{itemize}
            \item Machine Learning: \texttt{torch==2.9.0}, \texttt{torchrl==0.10.0}
            \item Others: \texttt{NumPy==2.2.6}, \texttt{gymnasium==0.29.1}
        \end{itemize}
        \item Simulator: \url{https://github.com/raphajaner/TUM_FTRT_simulator}
        \item Trajectory planning: \url{https://github.com/TUMFTM/global_racetrajectory_optimization}
    \end{itemize}

\clearpage

\section{Racetracks}\label{app:racetracks}

\begin{figure}[h]
    \centering
    \caption{Visualization of the synthetically-generated racetracks used in this work. The shape is inspired by real-world Roboracer competition, including various challenging curvy and high-speed sections. Each racetrack's width varies between a plus-or-minus $\SI{\pm20}{\percent}$ of the mean width, which is randomly selected between $\SI{1.5}{\meter}$ and $\SI{1.8}{\meter}$. The racetracks are named after famous cities from around the globe; however, there is no actual affiliation with the names. The last row shows the 5 racetracks only used for testing purposes and are not used for training.}
    \label{fig:all_racetracks}
    \includegraphics[width=0.8\linewidth]{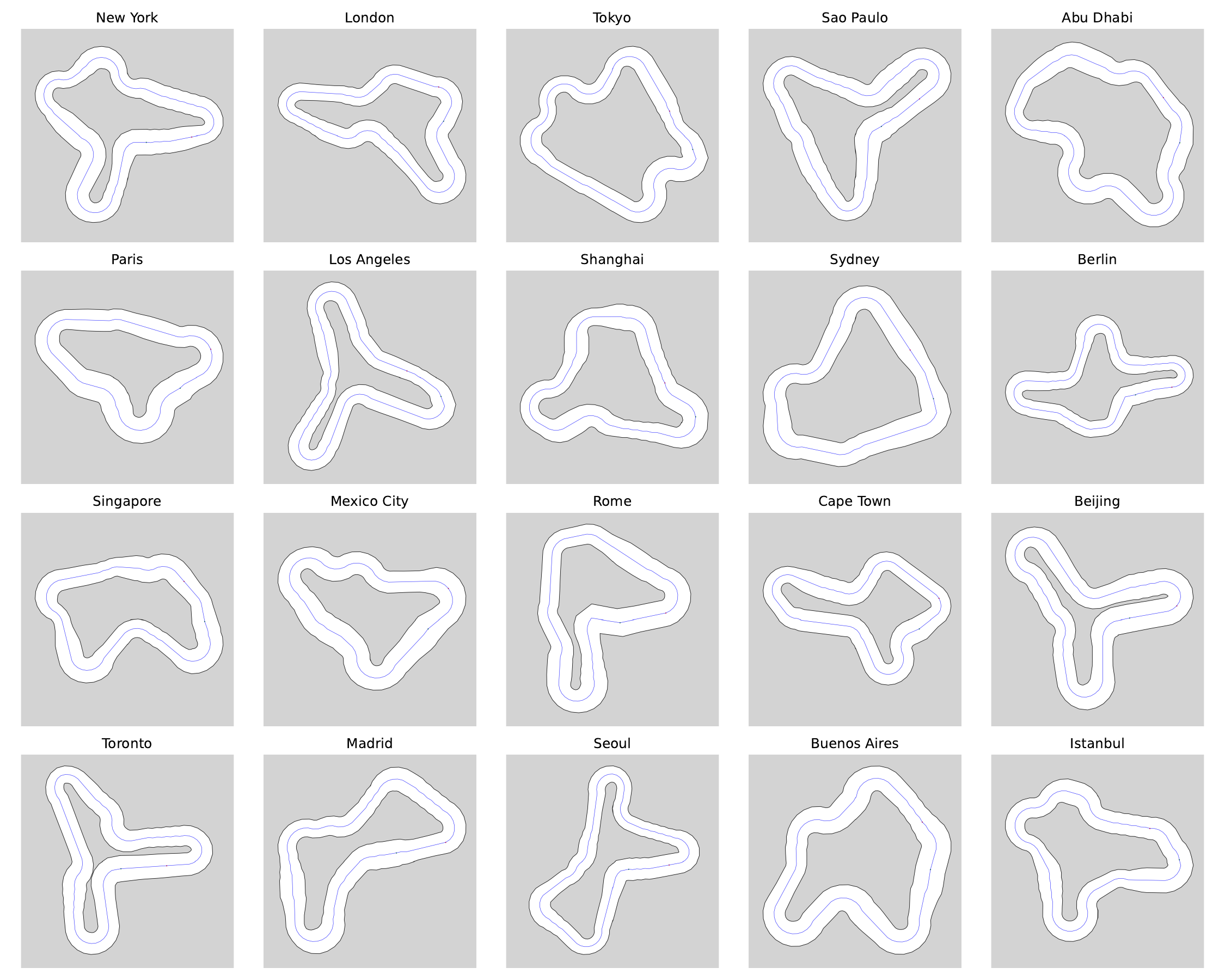}
\end{figure}

\begin{figure}[h]
    \centering
    \caption{Visualization of the \emph{Munich} racetrack, based on a SLAM mapping of our real-world testbed.}
    \label{fig:munich_racetrack}
    \includegraphics[width=0.35\linewidth]{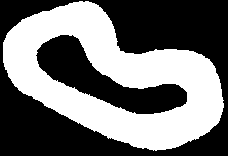}
\end{figure}

\clearpage

\section{Extended Material for Experiments}\label{app:extended_experiments}

    \subsection{Extended Race Results}
    
    \begin{figure}[!h]%
        \centering%
        \rotatebox{90}{%
        \begin{minipage}{0.9\textheight}
            \captionsetup{width=1.0\textwidth}
            \small
            \captionof{table}{Results for simulated results, showing all 15 training and 6 testing maps. The \emph{Munich} racetrack is an SLAM mapping of our real-world racetrack. A race finishes when \textit{5 laps} are completed. The best average results are highlighted in bold. The acronym BCD stands for BC+DRL, and DSSL is short for the DRL w/ SLL method. Note that due to numerical precision, average results can show zero collisions even when there are collisions on specific racetracks.}
            \label{app:tab:all_lap_results}
            \setlength{\tabcolsep}{0.8mm}%
            \begin{tabular}{p{0.2cm} p{1.8cm}|ccccccc|ccccccc|ccccccc}%
                \toprule%
                \multicolumn{2}{c|}{\textbf{Racetrack}}&\multicolumn{7}{c|}{\textbf{Total race time} $[\downarrow]$}&\multicolumn{7}{c|}{\textbf{Collisions per lap} $[\downarrow]$}&\multicolumn{7}{c}{\textbf{Max. speed} $[\uparrow]$}\\%
                &&FTG&Stanley&DRL&BCD&DSSL&RPL&\arpo{}&FTG&Stanley&DRL&BCD&DSSL&RPL&\arpo{}&FTG&Stanley&DRL&BCD&DSSL&RPL&\arpo{}\\%
                \midrule%
                \multirow{15}{*}{\rotatebox[origin=c]{90}{\textbf{Train}}}
                &Abu Dhabi&67.54&50.41&48.50&46.29&49.55&48.16&44.02&0.00&0.00&0.03&0.00&0.01&0.09&0.00&4.32&5.06&4.81&4.72&4.83&5.15&5.15\\%
                &Beijing&70.03&61.03&57.94&52.92&53.86&55.54&50.28&0.00&0.00&0.19&0.02&0.00&0.00&0.00&4.82&4.55&4.90&4.79&4.91&4.90&5.17\\%
                &Berlin&75.32&63.49&58.59&57.10&57.83&60.17&54.81&0.00&0.00&0.00&0.00&0.00&0.00&0.00&4.43&4.97&5.08&4.92&4.87&5.01&5.28\\%
                &Cape Town&72.64&57.25&51.88&51.19&52.39&51.95&49.15&0.00&0.00&0.00&0.00&0.00&0.00&0.00&4.98&4.69&5.01&4.88&4.93&4.94&5.28\\%
                &London&71.04&55.72&55.73&50.05&51.25&52.57&47.58&0.00&0.00&0.20&0.00&0.00&0.04&0.00&5.03&4.81&5.00&4.84&4.96&5.10&5.29\\%
                &Los Angeles&77.14&66.45&57.63&57.24&58.41&60.19&54.96&0.00&0.00&0.00&0.00&0.00&0.00&0.00&5.07&4.63&5.12&4.94&5.01&5.03&5.41\\%
                &Mexico City&55.78&41.15&37.51&37.54&38.15&37.78&35.53&0.04&0.00&0.00&0.00&0.00&0.00&0.00&4.78&5.21&5.10&4.92&5.02&5.37&5.50\\%
                &New York&71.88&63.08&54.49&54.00&55.09&56.17&51.95&0.00&0.00&0.00&0.00&0.00&0.00&0.00&4.58&4.45&5.08&4.95&5.02&4.89&5.34\\%
                &Paris&51.40&39.74&41.08&35.88&36.68&35.78&34.13&0.00&0.00&0.19&0.00&0.00&0.00&0.00&4.48&5.32&5.25&5.00&5.03&5.41&5.50\\%
                &Rome&73.55&52.90&48.20&48.09&48.98&48.99&45.61&0.12&0.00&0.00&0.00&0.00&0.00&0.00&5.16&5.37&5.23&4.96&5.05&5.42&5.59\\%
                &Sao Paulo&71.34&57.90&53.01&51.82&52.71&55.63&49.80&0.00&0.00&0.00&0.00&0.00&0.00&0.00&4.44&5.35&5.20&5.07&5.00&5.23&5.53\\%
                &Shanghai&64.94&48.34&46.88&46.46&46.60&46.63&44.89&0.00&0.00&0.01&0.00&0.00&0.00&0.00&4.32&5.08&5.19&5.00&5.07&5.30&5.55\\%
                &Singapore&66.27&50.25&47.16&46.47&46.77&48.12&44.58&0.00&0.00&0.00&0.00&0.00&0.00&0.00&4.23&5.43&5.13&4.97&5.06&4.94&5.50\\%
                &Sydney&55.95&44.52&40.18&40.16&40.90&40.43&38.37&0.00&0.00&0.00&0.00&0.00&0.00&0.00&4.94&4.98&5.23&4.91&4.98&5.38&5.44\\%
                &Tokyo&71.32&55.53&49.98&49.46&50.93&49.53&47.96&0.06&0.00&0.00&0.00&0.00&0.00&0.00&4.87&5.14&5.25&5.02&5.06&5.51&5.58\\%
                \midrule%
                &Average Train&67.74&53.85&49.92&48.31&49.34&49.84&\textbf{46.24}&0.01&\textbf{0.00}&0.04&0.00&0.00&0.01&\textbf{0.00}&4.70&5.00&5.10&4.93&4.99&5.17&\textbf{5.41}\\%
                \midrule%
                \multirow{6}{*}{\rotatebox[origin=c]{90}{\textbf{Test}}}
                &Buenos Aires&67.29&52.33&49.79&49.09&49.33&50.30&47.30&0.00&0.00&0.00&0.00&0.00&0.00&0.00&4.40&4.85&4.90&4.86&4.93&4.88&5.24\\%
                &Istanbul&69.62&52.27&48.15&53.01&49.22&86.42&45.86&0.00&0.00&0.00&0.19&0.00&1.24&0.00&4.48&5.32&4.95&4.82&4.97&5.30&5.29\\%
                &Madrid&66.91&53.51&49.66&49.09&49.60&50.87&47.28&0.00&0.00&0.00&0.00&0.00&0.00&0.00&4.97&4.46&5.03&4.88&4.89&4.79&5.32\\%
                &Seoul&77.78&65.95&61.57&59.96&60.77&65.19&57.76&0.00&0.00&0.00&0.00&0.00&0.08&0.00&4.94&5.03&5.21&5.00&5.06&5.10&5.43\\%
                &Toronto&76.31&65.20&59.63&57.92&58.64&62.67&55.58&0.00&0.00&0.00&0.00&0.00&0.00&0.00&5.18&4.60&5.16&4.95&5.04&4.95&5.35\\%
                &\emph{Munich}&38.10&32.77&41.87&33.60&30.15&32.71&28.91&0.00&0.00&0.41&0.14&0.00&0.01&0.00&3.41&4.29&4.69&4.70&4.70&4.46&4.92\\%
                \midrule%
                &Average Test&66.00&53.67&51.78&50.44&49.62&58.03&\textbf{47.11}&\textbf{0.00}&\textbf{0.00}&0.07&0.06&\textbf{0.00}&0.22&\textbf{0.00}&4.57&4.76&4.99&4.87&4.93&4.92&\textbf{5.26}\\%
                \bottomrule%
            \end{tabular}%
        \end{minipage}%
    }
    \end{figure}

    \subsection{Further Trajectories}

    \begin{figure*}[!h]
        \centering
        \begin{subfigure}[b]{0.32\textwidth}
            \centering
            \includegraphics[width=1.0\linewidth]{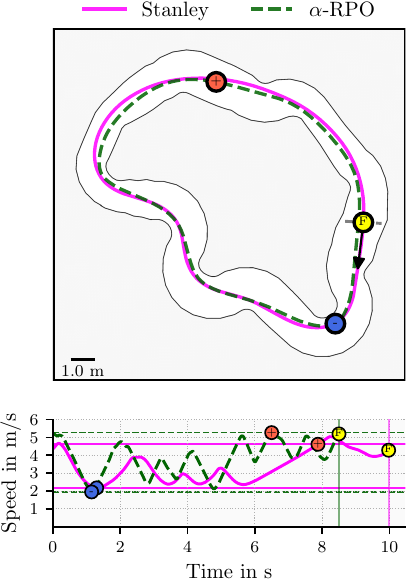}
            \caption{\arpo{} (green) against Stanley (magenta) on the Abu Dhabi racetrack (training).}
            \label{app:fig:trajectory_1_train}
        \end{subfigure}
        \hfill
        \begin{subfigure}[b]{0.32\textwidth}
            \centering
            \includegraphics[width=\linewidth]{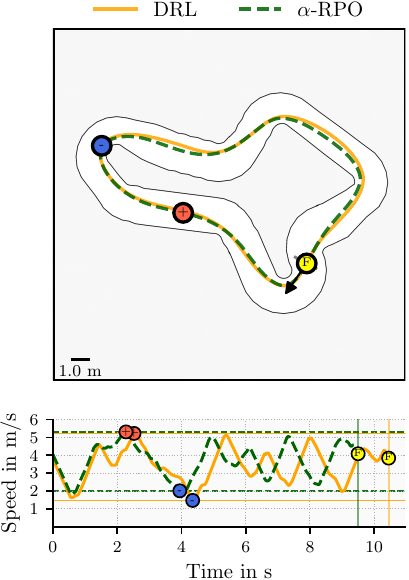}
            \caption{\arpo{} (green) against \gls*{drl} (orange) on the Cape Town racetrack (training).}
            \label{app:fig:trajectory_2_train}
        \end{subfigure}
        \hfill
        \begin{subfigure}[b]{0.32\textwidth}
            \centering
            \includegraphics[width=\linewidth]{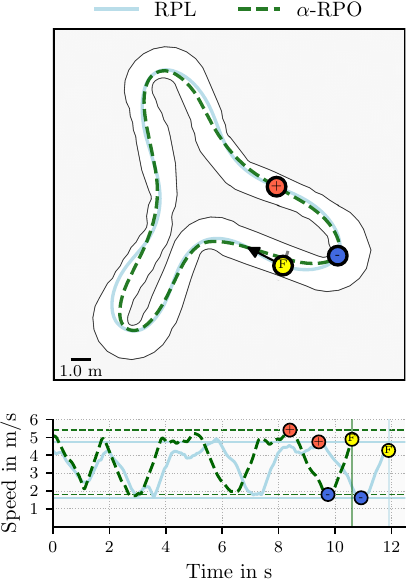}
            \caption{\arpo{} (green) against \gls*{rpl} (blue) on the Los Angeles racetrack (training).}
            \label{app:fig:trajectory_3_train}
        \end{subfigure}
        \caption{Qualitative comparison of trajectories and speed profiles on three \emph{training} racetracks of \arpo{} against Stanley (\textbf{left}), \gls*{drl} (\textbf{mid}), and \gls*{rpl} (\textbf{right}). A single flying lap is shown, where markers indicate a specific \gls*{poi} on the racetracks: $+$ marks a high-speed section (red), $-$ indicates the section with minimal speed (blue), while $\mathrm F$ is the finish line (yellow).
        }
        \label{app:fig:trajectory_analysis}
    \end{figure*}

    \begin{figure*}[!h]
        \centering
        \begin{subfigure}[b]{0.32\textwidth}
            \centering
            \includegraphics[width=1.0\linewidth]{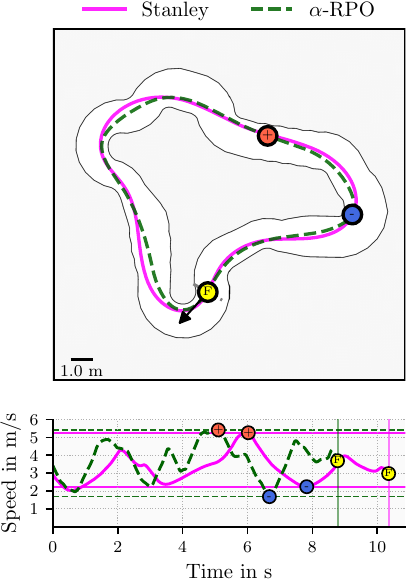}
            \caption{\arpo{} (green) against Stanley (magenta) on the Istanbul racetrack (training).}
            \label{app:fig:trajectory_1_test}
        \end{subfigure}
        \hfill
        \begin{subfigure}[b]{0.32\textwidth}
            \centering
            \includegraphics[width=\linewidth]{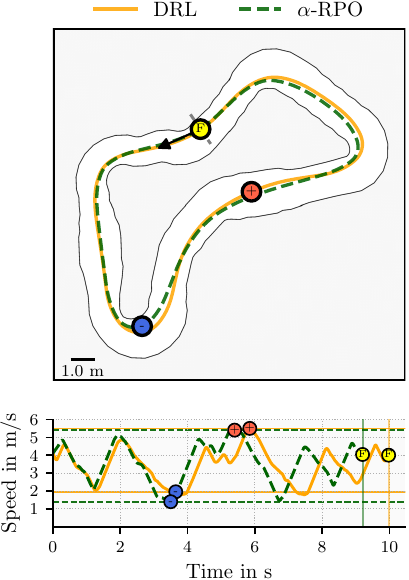}
            \caption{\arpo{} (green) against \gls*{drl} (orange) on the Madrid racetrack (training).}
            \label{app:fig:trajectory_2_test}
        \end{subfigure}
        \hfill
        \begin{subfigure}[b]{0.32\textwidth}
            \centering
            \includegraphics[width=\linewidth]{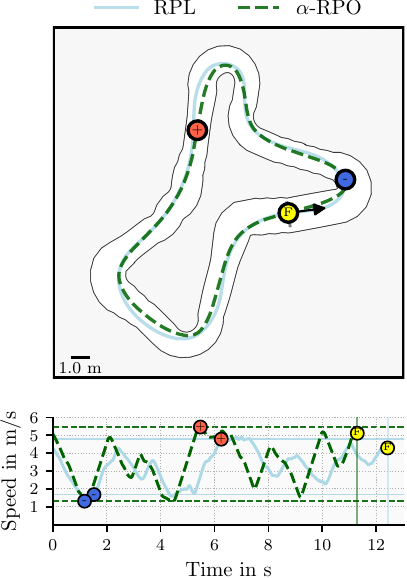}
            \caption{\arpo{} (green) against \gls*{rpl} (blue) on the Seoul racetrack (testing).}
            \label{app:fig:trajectory_3_test}
        \end{subfigure}
        \caption{Qualitative comparison of trajectories and speed profiles on three \emph{testing} racetracks of \arpo{} against Stanley (\textbf{left}), \gls*{drl} (\textbf{mid}), and \gls*{rpl} (\textbf{right}). A single flying lap is shown, where markers indicate a specific \gls*{poi} on the racetracks: $+$ marks a high-speed section (red), $-$ indicates the section with minimal speed (blue), while $\mathrm F$ is the finish line (yellow).
        } 
        \label{app:fig:trajectory_analysis2}
    \end{figure*}

\subsection{Further Real-World Trajectories}

\begin{figure*}[!h]
    \centering
    \begin{subfigure}[t]{0.49\textwidth}
      \centering
      \includegraphics[width=0.95\textwidth]{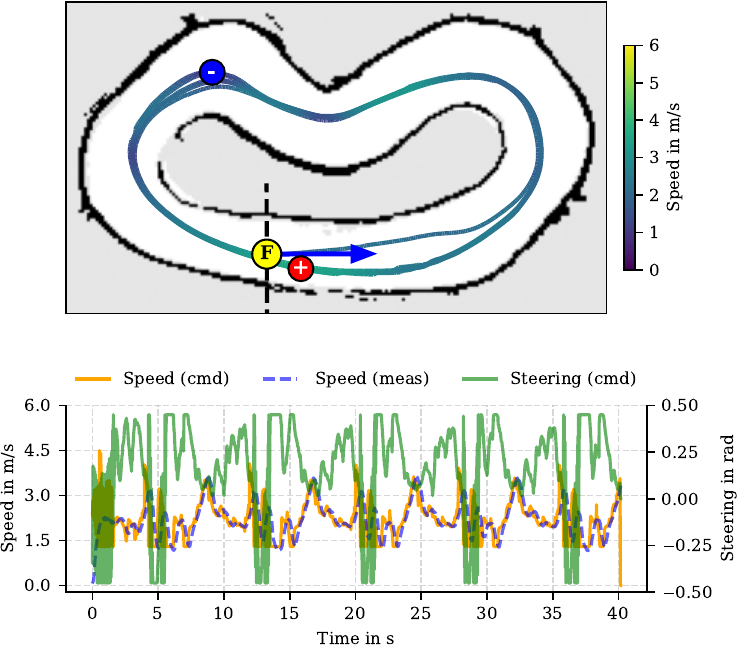}
      \caption{Real-world trajectories for \gls*{ftg} agent.}
      \label{app:fig:trajectory_ftg}
    \end{subfigure}
    \hfill
    \begin{subfigure}[t]{0.49\textwidth}
      \centering
      \includegraphics[width=0.95\textwidth]{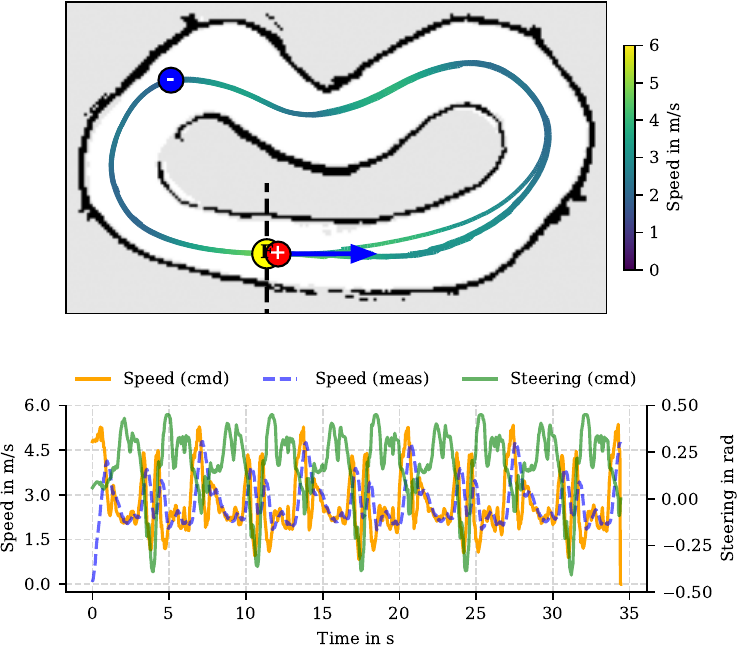}
      \caption{Real-world trajectories for \gls*{drl} agent.}
      \label{app:fig:trajectory_drl}
    \end{subfigure}
    \caption{Real-world trajectories for \gls*{ftg} (\textbf{left}) and \gls*{drl} (\textbf{right}) agents on the \emph{Munich} racetrack for 5 laps; current speed shown as color bar. Positions are estimated offline from \texttt{rosbag} data using SLAM. The agent's actions command (cmd) and measured speed (meas.) are also shown. Markers indicate a specific \gls*{poi} on the racetracks: $+$ marks a high-speed section (red), $-$ indicates the section with minimal speed (blue), while $\mathrm F$ is the finish line (yellow).}
\end{figure*}

    \subsection{Finetuning Comparison}
    
    \begin{figure*}[!h]
        \centering
        \begin{subfigure}[t]{0.49\textwidth}
          \centering
          \includegraphics[width=0.65\textwidth]{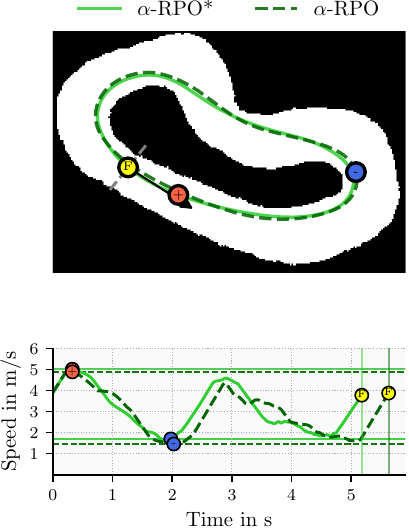}
          \caption{Comparison of simulated trajectory for \arpo{} and the corresponding finetuned \arpo{}\textsuperscript{*} agent.}
          \label{app:fig:trajectory_finetuned}
        \end{subfigure}
        \hfill
        \begin{subfigure}[t]{0.49\textwidth}
          \centering
          \includegraphics[width=0.95\textwidth]{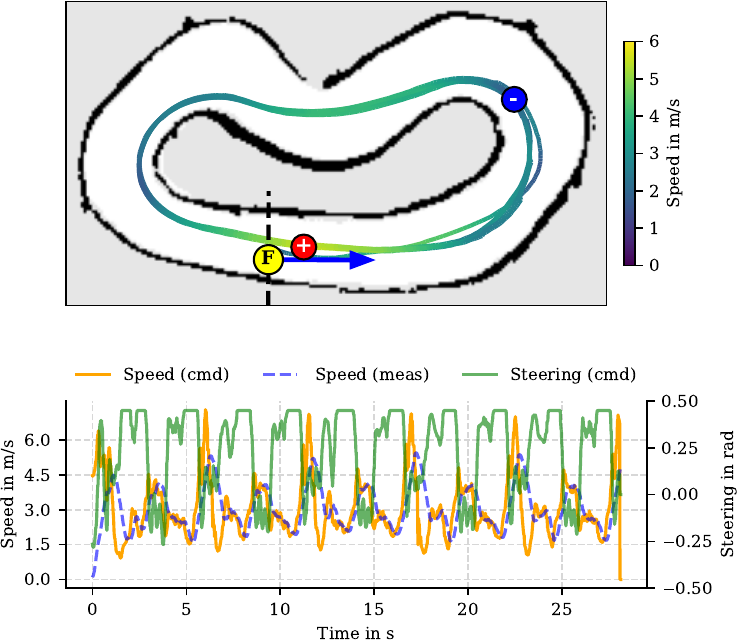}
          \caption{Real-world trajectories for finetuned \arpo{}\textsuperscript{*} agent.}
          \label{app:fig:trajectory_orig}
        \end{subfigure}
        \caption{Trajectories on the \emph{Munich} racetrack for 5 laps, comparing the \arpo{} agent to the finetuned \arpo{}\textsuperscript{*} version in simulation (\textbf{left}) and showing its real-world behavior (\textbf{right}). It can be seen that \arpo{}\textsuperscript{*} shows superior performance, particularly in the left-right section at the top, by cutting it straight with higher speed. The real-world trajectory shows the current speed as a color bar; positions are estimated offline from \texttt{rosbag} data using SLAM. The agent's actions command (cmd) and measured speed (meas.) are also shown. Markers indicate a specific \gls*{poi} on the racetracks: $+$ marks a high-speed section (red), $-$ indicates the section with minimal speed (blue), while $\mathrm F$ is the finish line (yellow).}
    \end{figure*}

\section{Neural Network Design}\label{app:nn_design}

    \subsection{Architecture}
    In this work, we employ a \gls*{dnn} architecture based on a shared encoder structure but with separate heads for the residual network $f_\theta$ and critic network $f_\phi$, totaling 344,357 trainable parameters; see Table~\ref{app:tab:model_arch} for parameter counts per layer.

    The architecture processes the downsampled LiDAR scan $L'_t \in [0,1]^{512}$ and the normalized vehicle state $s'_t$, which is a subset of $s_t$ excluding $L_t$, separately, and then fuses their embeddings.
    The \emph{LidarEncoder} processes $L'_t$ via a stack of five Conv1d layers, each with a kernel size of 3 and a stride of 2.
    To extract a compact feature representation from the resulting feature maps, we apply an AdaptiveSpatialSoftmax1d layer with $K=4$ keypoints; see Section~\ref{app:subsec:spatialsoftmax}.
    The resulting reduced feature vector is followed by a Linear projection, yielding the final LiDAR embedding $z_t \in \mathbb{R}^{128}$.
    Parallel to this, the vehicle state is encoded by the \emph{StateEncoder}, consisting of two stacked Linear layers, to produce the state embedding $h_t \in \mathbb{R}^{128}$.
    These embeddings, ${z}_t$ and $h_t$, are concatenated and refined within the \textit{FusionBlock} via LayerNorm normalization followed by a Linear layer to fuse them to the final embedding  ${z}'_t$.
    This fused feature vector serves as the input to three independent fully-connected heads that build the residual network $f_\theta$ and critic network $f_\phi$.
    The \emph{PolicyHead} and the \emph{LogStdHead} feature both a hidden dimension of 256, while the \emph{CriticHead} has a higher hidden dimension of 512 for increased descriptive capacity.

    \begin{figure}[!h]
        \centering
        \includegraphics[width=1.0\linewidth]{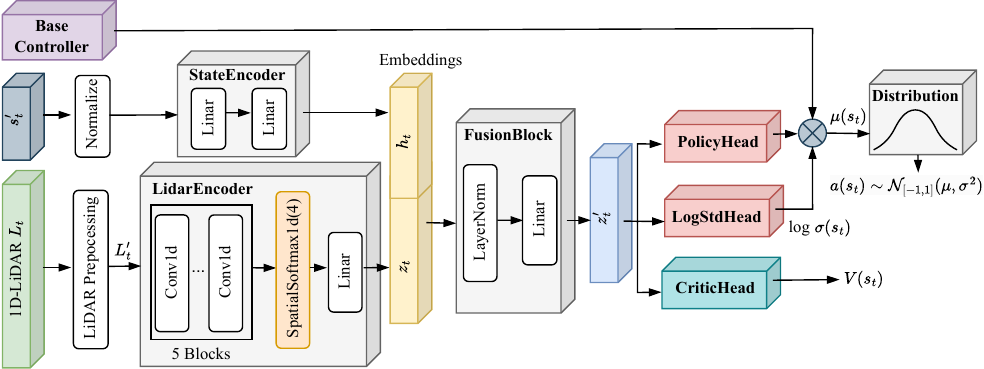}
        \caption{Proposed \gls*{dnn} architecture. The input to the \emph{LidarEncoder} is the stacked, processed LiDAR observations $L'_t$, and the \emph{StateEncoder} input is the normalized vehicle state $s'_t$, which is the subset of $s_t$ without $L_t$.
        The \emph{FusionBlock} outputs the embeddings $z'_t$, which are fed into the three independent heads: \emph{PolicyHead}, \emph{LogStdHead}, and \emph{CriticHead}.
        }
        \label{app:fig:network_architecture}
    \end{figure}

    \subsection{Adaptive SpatialSoftmax1d Layer}\label{app:subsec:spatialsoftmax}

    To efficiently transition from high-dimensional 1D sensor features to a compact state representation, we employ an adaptive \emph{SpatialSoftmax1d} layer.
    This mechanism adapts the deep spatial autoencoder architecture~\cite{finn2016deep} to 1D sequence data, serving as a learnable, geometry-preserving alternative to adaptive AveragePooling, which has recently shown great promise to improve generalization in visual \gls*{drl}~\cite{trumpp2025impoola}.
    Unlike standard pooling, which degrades or even removes spatial information, or flattening, which results in a large feature vector, this layer reduces the feature space while retaining topological structure.

    The adaptive SpatialSoftmax(K) layer compresses a feature map $X \in \mathbb{R}^{C \times L}$ of arbitrary length $L$ into a compact, fixed set of $K$ spatial keypoints for each of the $C$ channels.
    The input sequence is first partitioned into $K$ equal-length segments. For each channel $c$ and segment $k$, we compute a local probability distribution $P_{c,k}$ over the segment's domain by applying a softmax function
    \begin{equation}
        P_{c,k}(i) = \frac{\exp(X_{c,k,i} / T)}{\sum_{j} \exp(X_{c,k,j} / T)},
    \end{equation}
    where the temperature is typically set to $T=1$.
    We then compute the expected 1D position $\hat{x}_{c,k}$ within the segment's local coordinate frame
    \begin{equation}
        \hat{x}_{c,k} = \sum_{i} \xi_i \cdot P_{c,k}(i)
    \end{equation}
    where $\xi \in \mathbb{R}^{L/K}$ is a fixed linear coordinate grid linearly spaced in the interval $[-1, 1]$, representing the normalized position within any given segment.
    The computed scalar keypoints $\hat{x}_{c,k}$ for all channels $c \in \{1,\dots,C\}$ and segments $k \in \{1,\dots,K\}$ are concatenated to form the final output tensor $\hat{X} \in \mathbb{R}^{C \times K}$.
    As such, the SpatialSoftmax(K) layer leads to a reduction of the feature space by a factor of $L/K$.

    \subsection{Parameter Counts}

    \newcommand{\tree}[1]{\hspace{#1em}\ensuremath{\llcorner}}
    
    \begin{table}[!h]
    \centering
    \caption{Summary with parameters counts per layer of the \gls*{dnn} model.}
    \label{app:tab:model_arch}
        \begin{tabular}{l r r c c c}
            \toprule
            \textbf{Layer Type} & \textbf{Parameter Count} & \textbf{\%} & \textbf{Kernel} & \textbf{Input Size} & \textbf{Output Size} \\
            \midrule
            \tree{0} \textbf{Encoder} & -- & -- & -- & [B, 4, 1081] & [B, 128] \\
            \tree{2} LidarPreprocessor & -- & -- & -- & [B, 4, 1081] & [B, 4, 512] \\
            \tree{4} AvgPool1d & -- & -- & [2] & [B, 4, 1024] & [B, 4, 512] \\
            \tree{2} LidarEncoder & -- & -- & -- & [B, 4, 512] & [B, 128] \\
            \tree{4} Conv1d & 416 & 0.12\% & [3] & [B, 4, 512] & [B, 32, 256] \\
            \tree{4} SiLU & -- & -- & -- & [B, 32, 256] & [B, 32, 256] \\
            \tree{4} Conv1d & 6,208 & 1.80\% & [3] & [B, 32, 256] & [B, 64, 128] \\
            \tree{4} SiLU & -- & -- & -- & [B, 64, 128] & [B, 64, 128] \\
            \tree{4} Conv1d & 12,352 & 3.59\% & [3] & [B, 64, 128] & [B, 64, 64] \\
            \tree{4} SiLU & -- & -- & -- & [B, 64, 64] & [B, 64, 64] \\
            \tree{4} Conv1d & 24,704 & 7.17\% & [3] & [B, 64, 64] & [B, 128, 32] \\
            \tree{4} SiLU & -- & -- & -- & [B, 128, 32] & [B, 128, 32] \\
            \tree{4} Conv1d & 49,280 & 14.31\% & [3] & [B, 128, 32] & [B, 128, 16] \\
            \tree{4} AdaptSpatSoftmax1D & -- & -- & -- & [B, 128, 16] & [B, 128, 4] \\
            \tree{4} Flatten & -- & -- & -- & [B, 128, 4] & [B, 512] \\
            \tree{4} Linear & 65,664 & 19.07\% & -- & [B, 512] & [B, 128] \\
            \tree{2} StateEncoder & -- & -- & -- & [B, 16] & [B, 128] \\
            \tree{4} Linear & 2,176 & 0.63\% & -- & [B, 16] & [B, 128] \\
            \tree{4} SiLU & -- & -- & -- & [B, 128] & [B, 128] \\
            \tree{4} Linear & 16,512 & 4.80\% & -- & [B, 128] & [B, 128] \\
            \tree{2} FusionBlock & -- & -- & -- & [B, 128] & [B, 128] \\
            \tree{4} LayerNorm & 512 & 0.15\% & -- & [B, 256] & [B, 256] \\
            \tree{4} Linear & 32,896 & 9.55\% & -- & [B, 256] & [B, 128] \\
            \tree{0} \textbf{CriticHead} & -- & -- & -- & [B, 128] & [B, 1] \\
            \tree{4} Linear & 66,048 & 19.18\% & -- & [B, 128] & [B, 512] \\
            \tree{4} SiLU & -- & -- & -- & [B, 512] & [B, 512] \\
            \tree{4} Linear & 513 & 0.15\% & -- & [B, 512] & [B, 1] \\
            \tree{0} \textbf{PolicyHead} & -- & -- & -- & [B, 128] & [B, 2] \\
            \tree{4} Linear & 33,024 & 9.59\% & -- & [B, 128] & [B, 256] \\
            \tree{4} SiLU & -- & -- & -- & [B, 256] & [B, 256] \\
            \tree{4} Linear & 514 & 0.15\% & -- & [B, 256] & [B, 2] \\
            \tree{0} \textbf{LogStdHead} & -- & -- & -- & [B, 128] & [B, 2] \\
            \tree{4} Linear & 33,024 & 9.59\% & -- & [B, 128] & [B, 256] \\
            \tree{4} SiLU & -- & -- & -- & [B, 256] & [B, 256] \\
            \tree{4} Linear & 514 & 0.15\% & -- & [B, 256] & [B, 2] \\
            \midrule
            \multicolumn{2}{l}{\textbf{Total parameters:} 344,357} \\
            \bottomrule
        \end{tabular}%
    \end{table}

\end{document}